\documentclass[lettersize,journal]{IEEEtran}

\usepackage{cite}
\usepackage{amsmath,amssymb,amsfonts}
\usepackage{algorithmic}
\usepackage{graphicx}
\usepackage{textcomp}
\usepackage{xcolor}
\usepackage[linesnumbered,ruled,vlined]{algorithm2e}
\usepackage{hyperref}
\usepackage{booktabs}
\usepackage{enumitem}
\usepackage{makecell}
\usepackage{lettrine}
\usepackage{amsthm}
\usepackage{multirow}

\usepackage{xcolor}
\usepackage{listings}
\usepackage[most]{tcolorbox}

\lstdefinestyle{promptstyle}{
  basicstyle=\ttfamily\footnotesize,
  breaklines=true,
  breakatwhitespace=true,
  breakindent=0pt,
  breakautoindent=false,
  columns=fixed,
  keepspaces=false,      
  showstringspaces=false,
  frame=none
}

\newtcblisting{promptbox}{
  listing only,
  listing options={style=promptstyle},
  colback=gray!6,
  colframe=gray!6,
  boxrule=0pt,
  left=1.4mm,
  right=1.4mm,
  top=1.2mm,
  bottom=1.2mm,
  arc=0.8mm,
  breakable
}



\theoremstyle{remark}
\newtheorem{remark}{Remark}
\definecolor{deeprose}{HTML}{C71585}

\newcommand{\nan}[1]{\textcolor{black}{#1}}

\def\BibTeX{{\rm B\kern-.05em{\sc i\kern-.025em b}\kern-.08em
    T\kern-.1667em\lower.7ex\hbox{E}\kern-.125emX}}
    
\begin{document}

\title{Adaptive Obstacle-Aware Task Assignment and Planning for Heterogeneous Robot Teaming}

\author{Nan Li, Jiming Ren, Haris Miller, Samuel Coogan, Karen M. Feigh, and Ye Zhao
\thanks{The authors are with the Institute for Robotics and Intelligent Machines, Georgia Institute of Technology, Atlanta, GA 30332, USA. Email: \href{mailto:nan.li@gatech.edu}{\texttt{nan.li@gatech.edu}}}
\thanks{This work is sponsored by Lockheed Martin Corporation University Research program. The work is that of the authors and does not represent an official position of LMCO.}
}


\maketitle

\begin{abstract}
Multi-Agent Task Assignment and Planning (MATP) has attracted growing attention but remains challenging in terms of scalability, spatial reasoning, and adaptability in obstacle-rich environments.
To address these challenges, we propose OATH — Adaptive Obstacle-Aware Task Assignment and Planning for Heterogeneous Robot Teaming — 
which advances MATP by introducing a novel obstacle-aware strategy for task assignment. First, we develop an adaptive Halton sequence map, the first known application of Halton sampling with obstacle-aware adaptation in MATP, which adjusts sampling density based on obstacle distribution.
Second, we propose a cluster–auction–selection framework that integrates obstacle-aware clustering with weighted auctions and intra-cluster task selection. These mechanisms jointly enable effective coordination among heterogeneous robots while maintaining scalability and suboptimal allocation performance. In addition, our framework leverages an LLM to interpret human instructions and directly guide the planner in real time.  
We validate OATH in both NVIDIA Isaac Sim and real-world hardware experiments using TurtleBot platforms, demonstrating substantial improvements in task assignment quality, scalability, adaptability to dynamic changes, and overall execution performance compared to state-of-the-art MATP baselines. A project website is available at \href{https://llm-oath.github.io/}{https://llm-oath.github.io/}.
\end{abstract}

\vspace{0.6em}

\textbf{\small{\textit{Note to Practitioners}---}}\small{\textbf{Coordinating heterogeneous robot team in dynamic, obstacle-rich environments remains a major challenge in practical robotics.
This paper presents OATH, a hierarchical framework that integrates adaptive and obstacle-aware task assignment and planning for heterogeneous robot teams.
The practical contributions are threefold. First, we propose an adaptive Halton sequence map that automatically adjusts sampling density based on obstacle distribution. Combined with Dijkstra-based distances, this map provides realistic task-to-task costs, enabling robots to achieve high-quality task assignments while maintaining scalability. Second, we develop a heterogeneous cluster–auction–selection framework that reduces allocation complexity while respecting robot capacity and capability constraints, which are common in real industrial deployments. Third, we implement an LLM-guided interaction module that interprets natural language commands and supports real-time replanning during task execution, allowing human operators to dynamically adjust system behavior when needed.
\nan{The proposed framework is validated in both NVIDIA Isaac Sim and real-world hardware experiments using TurtleBot platforms. The experimental results show consistent allocation behavior, stable replanning, and reliable execution under realistic communication latency and sensing noise. These findings suggest that the proposed framework remains effective beyond simulation settings and can be implemented on physical multi-robot systems.}
}}

\vspace{0.8em}


\begin{figure}[t]
    \centering
    \includegraphics[width=\columnwidth]{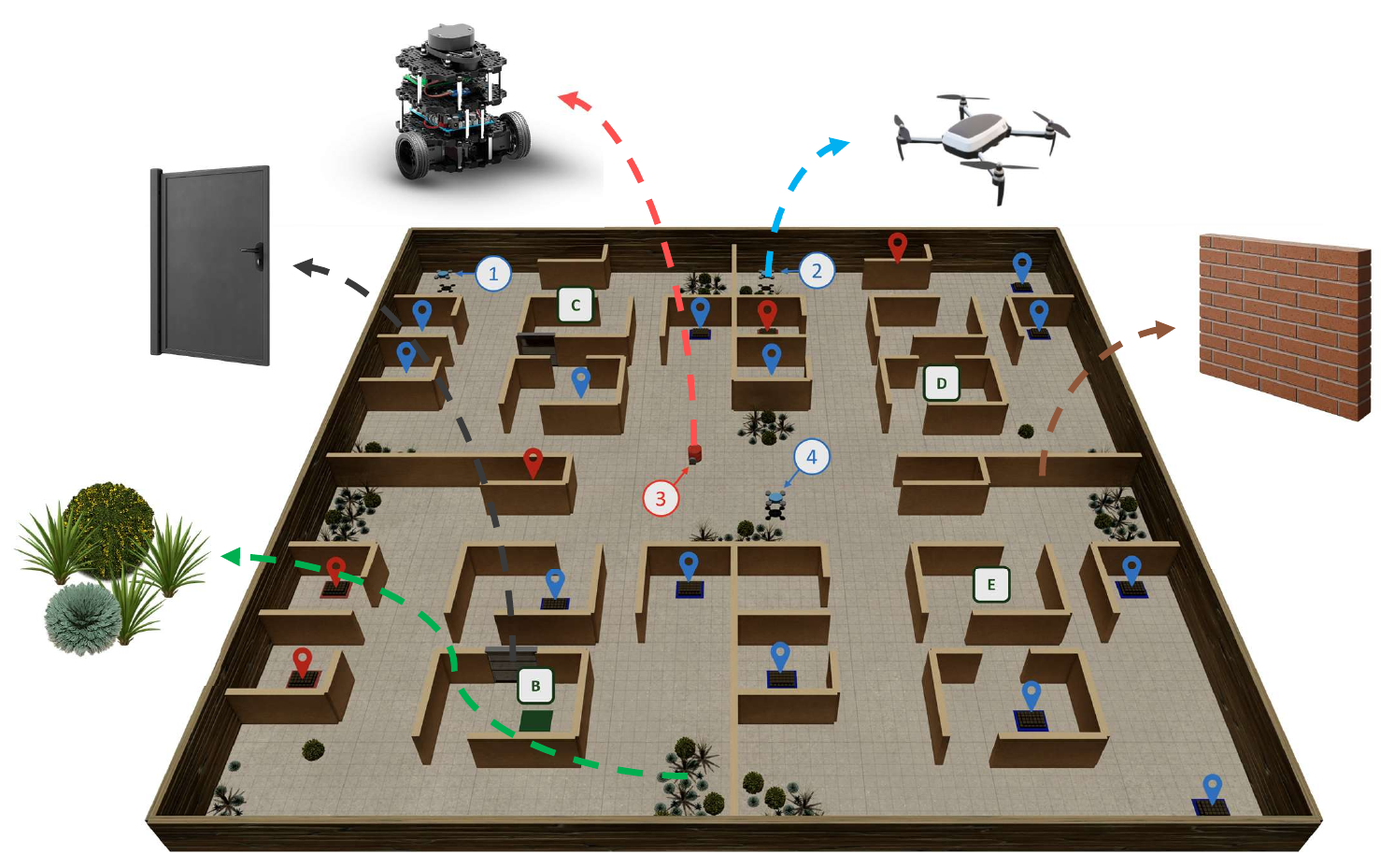}
    \caption{Problem setup in the Isaac simulation environment. A heterogeneous robot team of ground robots and drones is deployed in a maze-like environment. Two types of tasks are represented by blue and red markers, while delivery rooms are labeled \textbf{B–E}. The environment contains both known obstacles (walls) and initially unknown obstacles (iron gates and bushes).}
    \label{fig:isaac_env}
\end{figure}

\normalsize
\section{Introduction}

Multi-robot systems (MRSs) have attracted growing attention for their potential to enhance scalability, adaptability, and fault tolerance, yet they still present significant challenges in coordination, task assignment, and planning. Effective cooperation among multiple robots is essential for accomplishing complex objectives in domains such as agriculture, surveillance, search and rescue, and warehouse operations \cite{chakraa2023optimization, ju2022review, das2015distributed, mishra2020drone, cao2022leveraging}. A central challenge among many MRSs is the integration of task assignment and path planning to ensure coordinated and efficient operation \cite{chakraa2023optimization, salzman2020research, shamsah2025terrain, zhou2022reactive}.

In multi-robot scenarios (e.g., Fig~\ref{fig:isaac_env}), global missions are often decomposed into sub-tasks and distributed among individual agents. A widely studied formulation is the Multi-Agent Pickup and Delivery (MAPD) problem \cite{chen2021integrated}, which is common in real-world settings such as warehouse automation and logistics. Efficiently solving MAPD is essential for achieving scalability and coordination. MAPD generally involves two main components: task assignment and path planning.

For task assignment, two widely used approaches are auction-based methods \cite{choi2009consensus, de2022decentral, nunes2015multi, luo2015distributed} and optimization-based methods \cite{chakraa2023optimization}. Some works combine the two, for example by first clustering tasks with optimization techniques and then using auctions to allocate task clusters to robots \cite{padmanabhan2018multi, janati2016multi}. This hybrid method has proven effective for large-scale problems and improves both scalability and allocation quality. Our work follows this paradigm but introduces two key improvements: first, obstacle information is directly incorporated into the task assignment process; second, a hierarchical allocation framework is designed to enhance heterogeneity handling, efficiency, and scalability.

To achieve obstacle-aware task assignment, we develop an adaptive Halton sequence map. The Halton sequence is a classical sampling method that has been applied in robot motion and path planning to achieve more uniform coverage and smoother trajectories compared to grid-based maps \cite{velagic2014mobile, janson2018deterministic}. Prior works typically employ Halton or other quasi-random sequences for uniform roadmap construction or to improve sampling coverage in Probabilistic Roadmap Methods (PRM) and Rapidly-exploring Random Tree (RRT) style planners\cite{wang2020research}. However, these methods do not adapt the point distribution to the presence of obstacles. Recent studies on obstacle-aware sampling often rely on hard rejection (discarding points within obstacle regions) or potential-field-based biasing~\cite{tukan2022obstacle, ye2022real}. However, these approaches lack a principled way to adapt sampling density according to obstacle distribution.

To enable this principled mechanism, we extend Halton sampling into an adaptive obstacle-aware version, which to our knowledge is the first application of Halton with obstacle-aware adaptation in the Multi-Agent Task Assignment and Planning (MATP) setting. Specifically, we introduce a probabilistic acceptance function that adjusts point density according to obstacle proximity, which yields sparse sampling in open areas and dense sampling in cluttered regions. We then apply Dijkstra’s algorithm on the adaptive Halton sequence map to compute obstacle-feasible paths, which are subsequently used in the task assignment stage to explicitly account for obstacle constraints.

To improve efficiency and scalability in MATP, particularly for heterogeneous robot teams, we propose a hierarchical task assignment framework based on a cluster–auction–selection structure. This framework leverages the scalability of the cluster–auction approach while mitigating its suboptimality by incorporating obstacle-aware distances during clustering \cite{janati2016multi, you2023novel}. After cluster assignment, an additional intra-cluster optimization step is undertaken to select specific tasks, combining the strengths of auction-based and optimization-based methods. By restricting the optimization to a smaller intra-cluster scale, the framework also alleviates the computational overhead typically associated with global, large-scale  optimization\cite{chakraa2023optimization,patil2022algorithm}. Overall, this hierarchical design achieves a balance between optimality, scalability, and efficiency for heterogeneous robot teams.

While adaptive sampling addresses obstacle-aware task assignment and the hierarchical task allocation framework enhances efficiency and scalability, another critical aspect of practical multi-robot systems is the ability to incorporate real-time human instructions or intervention.

Recent advances in Large Language Models (LLMs), such as GPT-4 \cite{achiam2023gpt}, GPT-3.5 \cite{ye2023comprehensive}, LLaMa 3 \cite{grattafiori2024llama}, and Gemini 1.5 \cite{team2024gemini}, have demonstrated strong capabilities in natural language reasoning and generalization \cite{sun2024llm}. These abilities open new possibilities for human–robot interaction in multi-robot systems. Existing research has leveraged LLMs either as translators, converting natural language instructions into formal task representations~\cite{chen2024autotamp, liu2022lang2ltl, zhang2024lamma}, or as high-level planners for generating mission-level task structures~\cite{kannan2024smart, ahn2022can, huang2022language}. However, such approaches are typically limited to pre-mission planning and do not support online interaction during task execution. 

In real deployments, human operators often need to introduce new tasks, adjust priorities, or impose additional safety constraints during execution.
Existing research on human–robot interaction in multi-robot system primarily falls into three categories\cite{li2025large}: humans provide instructions only at the start of the mission, after which the robots execute tasks autonomously \cite{ahn2024vader}; humans give commands only after a task or robot failure occurs, when intervention is needed for recovery \cite{hunt2024conversational}; or interactive platforms allow humans to query robot status or ask clarifying questions, but not to modify behavior in real time\cite{lykov2023llm}.
To the best of our knowledge, there is still no MATP framework that allows humans to inject new instructions during execution and enables robots to immediately replan in response.

To address all above limitations, we propose \textbf{OATH}: Adaptive \textbf{O}bstacle-\textbf{A}ware \textbf{T}ask Assignment and Planning for \textbf{H}eterogeneous Robot Teaming, which achieves adaptivity in three aspects: obstacle-aware sampling, capability-aware task assignment, and real-time replanning based on human instructions.
Our main contributions are as follows:

\begin{itemize}
    \item We propose an \textbf{adaptive Halton sequence map}, where sampling density automatically adjusts according to obstacle distribution.
    
    \item We develop a \textbf{hierarchical heterogeneous allocation framework} based on a cluster–auction–task selection structure, which generalizes to any number of task types and reduces allocation complexity while maintaining capacity and capability constraints.
    
    \item We present a \textbf{fully integrated online pipeline} with LLM-guided interaction for multi-robot system. It interprets natural language inputs and supports real-time replanning in response to human instructions.
    
    \item \nan{We validate the OATH framework through extensive simulation case studies and real-world hardware experiments, showing substantial improvements in task assignment quality, scalability, adaptability to dynamic changes, and overall execution performance compared to state-of-the-art MATP baselines.}
\end{itemize}

\section{Related Work}
\subsection{Task Assignment}

Multi-robot Systems (MRSs) are prevalent in applications such as search and rescue, pickup and delivery, and target detection. Multi-robot task assignment focuses on assigning tasks to robot teams to optimize metrics like mission completion time. These assignment methods are typically categorized into centralized and decentralized approaches.

\textbf{Centralized algorithms} provide near-optimal solutions but often rely on stable communication and global information. Early work explored metaheuristics such as genetic algorithms for UAV task assignment with precedence and timing constraints \cite{shima2006multiple}, while later studies focused on improving scalability through approaches like multi-objective particle swarm optimization for coalition formation in disaster response \cite{mouradian2017coalition} and federated optimization for dynamic ridesharing \cite{simonetto2019real}. Other efforts have developed heuristic methods for cooperative scheduling under precedence constraints \cite{bischoff2020multi}.

Despite these advances, centralized methods require reliable communication. To address this, Otte et al. \cite{otte2020auctions} proposed auction-based techniques that maintain performance under limited connectivity, serving as a middle ground between centralized and distributed planning.

\textbf{Decentralized approaches} rely on local information, offering better scalability and robustness in dynamic or communication-limited settings, though often at the cost of optimality. Auction-based methods are widely studied, including consensus-based bidding frameworks \cite{choi2009consensus}, extensions with conflict resolution mechanisms \cite{lindsay2021sequential}, and group-based auctions that handle capacity and time constraints \cite{bai2022group}. Other strategies focus on task updates in dynamic environments, such as greedy reallocation based on marginal cost changes \cite{zhao2015heuristic}.

\textbf{Learning-based approaches} have been increasingly adopted to enhance adaptability and scalability in multi-robot task assignment in recent research. Deep reinforcement learning (DRL) and graph neural network (GNN)-based methods enable agents to learn allocation strategies directly from experience, allowing generalization beyond handcrafted heuristics. Representative works include attention-inspired DRL frameworks for large-scale warehouse allocation~\cite{agrawal2022rtaw}, graph-based multi-agent reinforcement learning for heterogeneous teams~\cite{ratnabala2025magnnet}, and neural combinatorial optimization approaches using attention or GNN architectures~\cite{zhang2024scalable}.

While these learning-based methods demonstrate impressive adaptability, they often struggle to guarantee conflict-free task assignment or consistent convergence, occasionally yielding lower success rates compared to model-based or auction-based methods. For example, learning-enhanced consensus schemes such as GCN-augmented Consensus-Based Bundle Algorithm (CBBA)~\cite{chekakta2024towards} combine neural scoring with classical distributed auctions, but still rely on heuristic conflict resolution.

In our framework, we combine centralized and decentralized task assignment. A centralized cluster planner first performs task clustering based on global information. Each robot is then assigned to a specific cluster, within which it independently selects tasks using a decentralized strategy. This hybrid structure allows the system to retain the global efficiency of centralized planning while benefiting from the flexibility and fault tolerance of decentralized execution.

\subsection{Multi-Agent Pickup and Delivery}

The Multi-Agent Pickup and Delivery (MAPD) problem involves a team of robots executing a continuous stream of pickup and delivery tasks. Each task requires transporting an object from a pickup to a delivery location. Challenges include real-time task assignment, collision-free path planning, and constraints such as capacity, heterogeneity, and time windows. Existing methods can be divided into decoupled and coupled approaches.

\textbf{Decoupled methods} separate assignment and path planning to improve scalability. Recently, the Multi-Goal MAPD (MG-MAPD) framework \cite{xu2022multi} generalized the MAPD configuration to allow robots to carry and deliver multiple items across multiple goals, where task grouping and assignment are done offline, followed by prioritized path planning. Though generally suboptimal, such decoupled strategies support large-scale and online applications with lower computational demands.

\textbf{Coupled methods} solve task assignment and path planning jointly. CBS-TA \cite{honig2018conflict} modifies Conflict-Based Search to achieve optimal coordination but scales poorly. To improve performance, Chen et al. \cite{chen2021integrated} proposed a marginal-cost based assignment strategy with meta-heuristic refinements, which outperforms greedy baselines under capacity constraints. These approaches improve global solution quality but are computationally intensive. Ma et al. \cite{ma2017lifelong} introduced Token Passing with Task Swaps (TPTS), a decentralized online method that assigns tasks greedily and uses Cooperative A* for path planning, which can scale the problems up but cannot promise the solution quality.

Our proposed method follows a decoupled framework, where task assignment and path planning are solved separately. However, unlike standard decoupled approaches, we incorporate environmental information and obstacle layouts during task clustering and allocation. This ensures that the resulting task assignments are consistent with actual path feasibility, reducing mismatch between planning stages. As a result, the approach preserves solution quality while maintaining the scalability required for large-scale scenarios.

\begin{figure*}[t]
    \centering
    \includegraphics[width=0.85\textwidth]{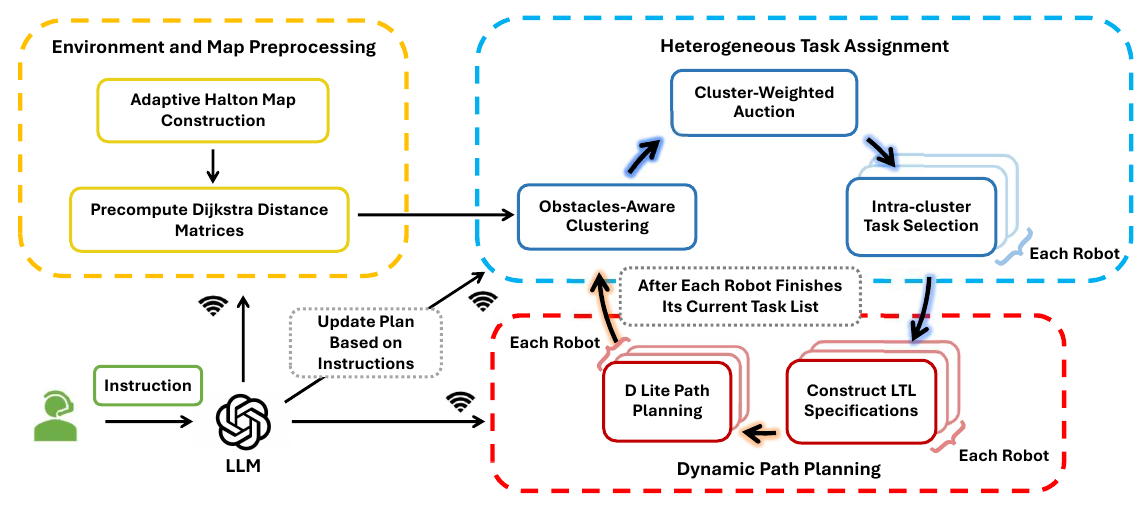}
    \caption{Overview of the OATH framework: the planning module is composed of three parts — environment and map preprocessing, task assignment, and path planning. Task assignment and path planning operate in a closed loop, where robot plans are iteratively updated as tasks are completed. At the high level, an LLM continuously interprets human instructions, providing semantic guidance to ensure obstacle-aware and adaptive task assignment throughout the process.}
    \label{fig:oath_framework}
\end{figure*}

\subsection{LLM-based Robot Planning}

Recent advances in Large Language Models (LLMs) have provided new possibilities for robot planning by enabling flexible interpretation of natural language and high-level reasoning. Current work that integrates LLMs into robot planning can be grouped into two main paradigms: \textit{LLMs as planners} and \textit{LLMs as translators} \cite{zhao2024survey}.

\textbf{LLMs as planners} use LLMs directly to generate task or motion plans from natural language instructions. Early examples include Naive Task Planning \cite{huang2022language}, which outputs a full sub-task sequence without validation, and SayCan \cite{ahn2022can}, which improves feasibility by scoring actions with a likelihood model. Later systems such as Text2Motion \cite{lin2023text2motion} and SMART-LLM \cite{kannan2024smart} combine LLM reasoning with feasibility checks or multi-agent coordination. While these works highlight the potential of LLMs for high-level planning, their performance is often constrained by limited spatial grounding and execution consistency.

\textbf{LLMs as translators}, in contrast, aim to convert natural language into structured task representations that can be solved by established model-based planners. Examples include translations into PDDL \cite{xie2023translating, liu2023llm+, zhang2024lamma}, temporal logics such as LTL or STL \cite{liu2022lang2ltl, chen2024autotamp}, and behavior trees \cite{styrud2024automatic}. This paradigm offers greater stability, modularity, and compatibility with existing robotics pipelines, since the heavy lifting of planning is delegated to well-studied solvers. Recent studies have shown that translator-based pipelines often outperform planner-based ones, particularly in spatially complex environments \cite{chen2024autotamp}. Furthermore, vision-language models (VLMs) have been introduced to enhance perceptual grounding and bridge the gap between high-level specifications and real-world observations \cite{liu2024lang2ltl}.

In our proposed method, we follow the translator paradigm but extend it to real-time use. Instead of translating instructions only once during planning, we integrate the LLM as a persistent interpreter throughout execution. When receiving a command like “There is a new task in room B; avoid the bushes near the door,” the LLM identifies actionable content, constraints, and dispatches relevant information to the appropriate module. This design enables dynamic instruction handling and modular system response.

 \begin{figure*}[t]
    \centering
    \includegraphics[width=\textwidth]{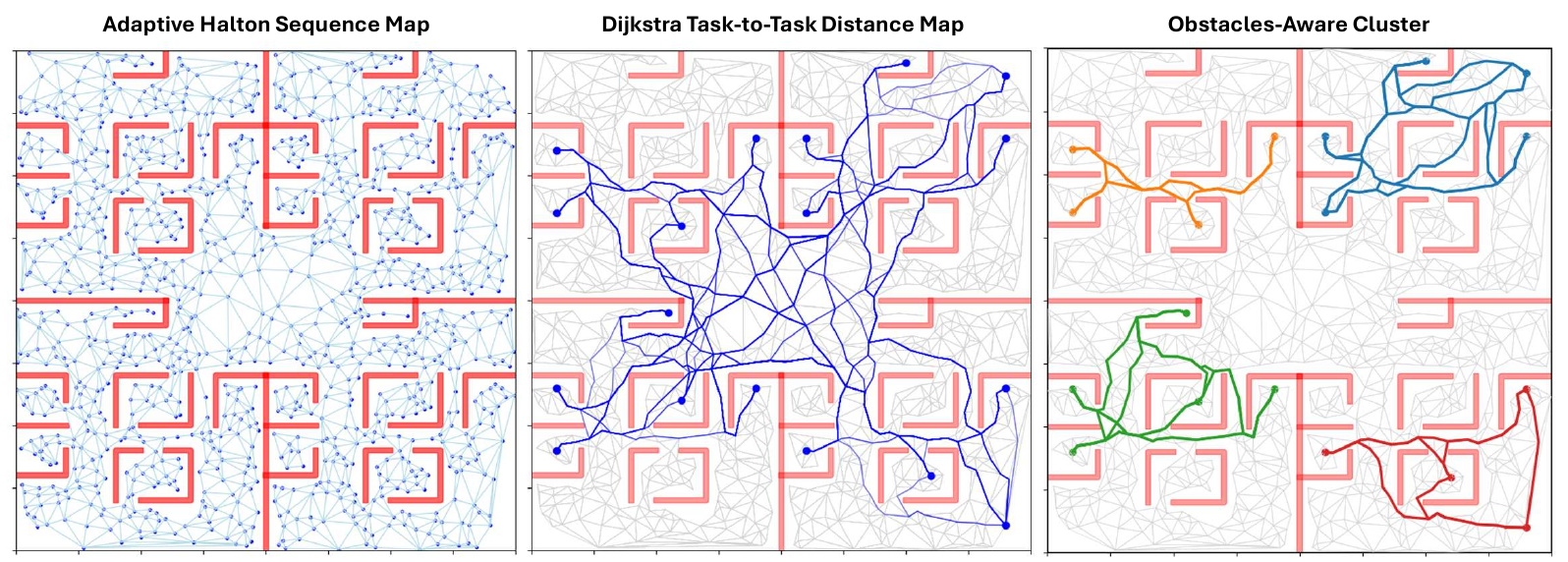}
    \caption{Spatial modeling and task clustering process. The left figure shows the construction of the adaptive Halton sequence, generating obstacle-aware sampling points. The middle figure presents the Dijkstra result computed over the Halton map, producing a task-to-task distance matrix. The right figure visualizes the final clustering result: four clusters are generated using agglomerative clustering \cite{mullner2011modern} based on the distance matrix, where each color represents one cluster. Notably, the clustering respects the environmental complexity, avoiding grouping tasks across walls or obstacle-dense regions.}
    \label{fig:spatial_pipeline}
\end{figure*}

\section{Problem Formulation}

We consider an MAPD problem involving multiple classes of heterogeneous robots and multiple types of tasks (see Fig.~\ref{fig:isaac_env}).
Let $\mathcal{P} = \{1, 2, \dots, P\}$ denote the set of tasks, and $\mathcal{R} = \{1, 2, \dots, R\}$ denote the set of robots.
Each task $i \in \mathcal{P}$ is associated with a pickup location $o_i$, a delivery location $d_i$, and a task type $\tau_i \in \mathcal{T}$, where $\mathcal{T}$ denotes the set of all task types and $|\mathcal{T}|$ is the total number of distinct task types.

Each robot $r \in \mathcal{R}$ is characterized by a capability vector $\boldsymbol U_r \in \{0,1\}^{|\mathcal{T}|}$ (or more generally $\boldsymbol U_r \in \mathbb{R}_{\ge 0}^{|\mathcal{T}|}$), where $(\boldsymbol U_r)_\tau = 1$ indicates that robot $r$ can perform task type $\tau \in \mathcal{T}$, and $(\boldsymbol U_r)_\tau = 0$ otherwise.
This representation captures a wide range of heterogeneous team configurations by allowing robots to have distinct or overlapping capabilities across different task types.

In the meantime, each robot $r \in \mathcal{R}$ possesses a capacity $Q_r$ indicating the maximum number of tasks it can carry simultaneously. The planned execution route for robot $r$ is denoted by an ordered sequence of pickup and delivery locations, $\pi_r = [\pi_r^1, \pi_r^2, \dots]$. Specifically, $\pi_r$ includes all pickup and delivery points assigned to robot $r$ in the current allocation round. For example, if robot $r$ is assigned to pick up items at locations $a$ and $b$ and deliver them to location $c$, then $\pi_r = [a, b, c]$ contains $a$, $b$ and $c$ in the order determined by the task assignment.

We define a binary assignment variable $\mu_{i,r} \in \{0,1\}$, which equals 1 if task $i$ is assigned to robot $r$, and 0 otherwise. Let $a(j)$ represent the arrival time of a robot at location $j$.
The function $q_r(t)$ denotes the number of tasks being carried by robot $r$ at time $t$. The travel time between any two locations $i$ and $j$ is denoted by $t(i,j)$.

\vspace{0.5em}
\noindent \textbf{Problem 1.}
Given a set of pickup and delivery tasks $\mathcal{P}$ and a heterogeneous robot team $\mathcal{R}$, the objective is to assign each task $i \in \mathcal{P}$ to a robot $r \in \mathcal{R}$ and plan a route $\pi_r$ for each robot to find an optimal solution, i.e., minimize the total travel cost.

The travel cost for robot $r$ is defined as the total travel time along its planned route, considering its motion and capability constraints:
\begin{equation}
    \sum_{j=1}^{|\pi_r| - 1} t(\pi_r^j, \pi_r^{j+1})
\end{equation}
%
The overall objective is to minimize the total travel cost of all robots:
\begin{subequations}\label{eq:vrp_opt}
\begin{alignat}{2}
\min \quad & \sum_{r \in \mathcal{R}}\sum_{j=1}^{|\pi_r|-1} t(\pi_r^j, \pi_r^{j+1}) \label{eq:vrp_obj} \\
\text{s.t.}\quad 
& \mathcal{P}_r \cap \mathcal{P}_{r'} = \emptyset, && \forall r, r' \in \mathcal{R},\ r \neq r', \label{eq:vrp_disjoint} \\
& \bigcup_{r \in \mathcal{R}} \mathcal{P}_r = \mathcal{P}, && \label{eq:vrp_cover} \\
& q_r(t) \leq Q_r, && \forall r \in \mathcal{R},\ \forall t, \label{eq:vrp_capacity} \\
& \mu_{i,r} = 0, && \text{if } (\boldsymbol U_r)_{\tau_i} = 0, \ \forall i \in \mathcal{P},\ \forall r \in \mathcal{R}, \label{eq:vrp_type_general} \\
& a(o_i) < a(d_i), && \forall i \in \mathcal{P}, \label{eq:vrp_precedence} \\
& \mu_{ir} = 1 \Rightarrow o_i, d_i \in \pi_r, && \forall i \in \mathcal{P},\ \forall r \in \mathcal{R}. \label{eq:vrp_assignment}
\end{alignat}
\end{subequations}
Constraint~\eqref{eq:vrp_disjoint} demands that each task is assigned to exactly one robot, and Constraint~\eqref{eq:vrp_cover} guarantees full task coverage. 
Constraint~\eqref{eq:vrp_capacity} enforces the capacity limit of each robot. 
Constraint~\eqref{eq:vrp_type_general} generalizes the type compatibility rule: a robot $r$ can only be assigned tasks whose type $\tau_i$ satisfies $(\boldsymbol U_r)_{\tau_i}=1$, i.e., the robot is capable of executing that task type.
Constraint~\eqref{eq:vrp_precedence} imposes pickup-before-delivery ordering for each task, and Constraint~\eqref{eq:vrp_assignment} requires that both pickup and delivery locations of any assigned task appear in the corresponding robot’s route.

Due to the NP-hard nature \cite{chakraa2023optimization} of Problem 1, finding an optimal solution is computationally expensive, particularly for large-scale scenarios. To address this challenge, we scaffold an efficient and modular planning framework, as illustrated in Fig.~\ref{fig:oath_framework}, which incorporates a high-level planner for task assignment and low-level planner for path routing. In addition, we integrate it with a human-in-the-loop interface powered by an LLM for enhanced adaptability and interaction.

\section{Methods}
\label{sec:task_assignment}

Many existing approaches for MATP perform task grouping prior to assignment, but typically rely on geometric proximity (e.g., Euclidean distance) without explicitly accounting for environmental obstacles \cite{janati2016multi, yuan2024multi, you2023novel}. In cluttered or complex environments, this often leads to inaccurate task groupings and, consequently, suboptimal task assignments and inefficient path planning.

To address this limitation, we introduce obstacle awareness into the task assignment pipeline through two key strategies. First, we construct an adaptive Halton sequence map that generates sampling points whose density varies according to the obstacle distribution. Second, we apply a Dijkstra algorithm to compute a task-to-task distance matrix that accurately reflects traversability under obstacle constraints.

Based on this environment representation, our obstacle-aware task assignment and planning framework operates fully online and adapts to heterogeneous robot teams and dynamic environments. The process begins with a cluster-based auction mechanism that assigns groups of tasks to individual robots. Each robot then performs intra-cluster task selection with capacity constraints and generates a suboptimal global path to execute its assigned tasks. Upon completing a task set, the robot re-enters the assignment and planning cycle. This iterative process continues until all tasks are completed.

An overview of the full pipeline is illustrated in Fig.~\ref{fig:oath_framework}, and the major components are detailed in the following subsections. \nan{We also provide a formal asymptotic complexity analysis to characterize how the computational cost scales with problem size in Appendix~\ref{app:complexity}.}

\subsection{Environment and Map Preprocessing}
The environment and map preprocessing stage consists of two main components: (1)  adaptive Halton sequence map construction, and (2) Dijkstra distance matrix computation.

\subsubsection{Adaptive Halton Sequence Map}
To provide a flexible and obstacle-aware spatial foundation for both task assignment and path planning, we adopt an \textit{adaptive Halton sequence map} instead of a conventional grid-based representation. 

A Halton sequence $\{h_i\}_{i=1}^N \subset [0,1]^{\epsilon}$ is an $\epsilon$-dimensional low-discrepancy sequence built from pairwise coprime bases $b_1,\dots,b_{\epsilon}$.
For each point index $i\in\{1,\dots,N\}$, write
\[
  h_i=\bigl(h_i^{(1)},\dots,h_i^{(\epsilon)}\bigr),
\]
where $h_i^{(j)}$ is the $j$-th coordinate of the $i$-th point, with $j\in\{1,\dots,\epsilon\}$.
For each $j$, this coordinate is given by the radical inverse in base $b_j$:

\begin{equation}
    h_i^{(j)} = \phi_{b_j}(i) = \sum_{k=0}^{L} \frac{\alpha_k}{b_j^{\,k+1}}, \quad i = \sum_{k=0}^{L} \alpha_k b_j^k
\end{equation}
where $\alpha_k$ are the digits of $i$ in base $b_j$. The resulting sequence fills the space $[0,1]^\epsilon$ more evenly than pseudo-random points.

Building on the Halton sequence, we construct an adaptive Halton sequence map by introducing an acceptance probability function that determines whether a candidate point is sampled at a given location. The acceptance probability for a candidate point is defined as:
\begin{equation}
P_{\text{accept}}(\delta) =
\begin{cases}
0, & \delta < \delta_{\min} \\
\beta + (1 - \beta) \exp\left(-\frac{(\delta - \delta_{\text{opt}})^2}{2\sigma^2} \right), & \delta \geq \delta_{\min}
\end{cases}
\label{eq:accept_prob}
\end{equation}
where \( \delta \) is the distance from the candidate point to the nearest obstacle, \( \delta_{\min} \) is the minimum safe distance, and \( \delta_{\text{opt}} \) is the optimal distance for sampling where the acceptance probability is highest. The parameter \( \sigma \) controls the width of the Gaussian distribution, and \( \beta \in [0,1] \) represents the minimum acceptance probability (i.e., the baseline floor value) in moderately constrained areas.

For each Halton-generated candidate point, we draw a random variable \( u \in [0,1] \). The point is accepted if it satisfies the condition \( u < P_{\text{accept}}(d) \). This process results in a non-uniform, obstacle-aware distribution of sampling points that is used consistently across task preprocessing and path planning.

The left image in Fig.~\ref{fig:spatial_pipeline} shows the result of this adaptive sampling process. Compared to a regular grid that produces a fixed lattice, the adaptive Halton sequence provides non-uniform yet well-distributed sampling, offering smoother spatial coverage and improved connectivity that naturally forms a triangular navigation structure rather than axis-aligned grid edges.
When applied in our setting, these properties yield three key advantages over grid maps with the same number of sampled points: (1) the adaptive Halton map offers more accessible entries to narrow passages, increasing the likelihood of finding alternate routes in multi-robot planning; (2) paths generated on the Halton map are typically smoother and shorter as they avoid the fixed $90^{\circ}$ or $45^{\circ}$ turns inherent in grid-based maps; and (3) the adaptive Halton sampling can dynamically adjust point density to local obstacle distributions, a flexibility that fixed grid representations lack.

\begin{remark}[Practical Parameter Tuning]
    \nan{
    The parameters in the probabilistic acceptance function of Eq.~\ref{eq:accept_prob} are tunable and have intuitive geometric interpretations.
    The threshold $\delta_{\min}$ enforces a safety margin from obstacles, while $\delta_{\text{opt}}$ specifies the distance at which sampling is most informative for navigation.
    The parameters $\sigma$ and $\beta$ control the smoothness of density decay and the minimum sampling density in open regions, respectively.
    In practice, these parameters are adjusted based on external factors such as the map scale, obstacle density, and robot footprint.
    A typical tuning procedure is as follows.
    First, $\delta_{\min}$ is set according to the robot size and safety margin to ensure collision-free sampling.
    Next, $\delta_{\text{opt}}$ is chosen slightly larger than $\delta_{\min}$ to promote denser sampling near obstacle boundaries, which are critical for feasible navigation.
    The parameter $\sigma$ is then used to regulate how broadly this dense sampling region extends, with larger values producing smoother density transitions and smaller values emphasizing narrow passages.
    Finally, $\beta$ is selected to guarantee a minimum sampling density in open regions.
    The tuning objective is to ensure that all task locations remain mutually reachable in the resulting roadmap, while narrow passages are sufficiently sampled to admit multiple feasible traversal paths.
    Once this connectivity criterion is satisfied, further adjustment of parameters has limited impact on downstream task allocation performance.
    We empirically evaluated the effect of varying $\delta_{\text{opt}}$, $\sigma$, and $\beta$ within practical ranges.
    Across a wide range of parameter combinations that satisfy the above connectivity objective, we observe similar roadmap connectivity and comparable task allocation performance, which indicates that the method is not sensitive to precise parameter values.}
\end{remark}

\subsubsection{Dijkstra-based Distance Matrix}
\label{sec:dijkstra_matric}

In our framework, we apply a Dijkstra algorithm to the previously constructed adaptive Halton sequence map. Specifically, each task location, both pickup and delivery points, is treated as a source node, and shortest paths to all other task locations are calculated in advance. This produces a task-to-task distance matrix 
\begin{equation}
\begin{split}
&\mathcal{M} = \big[ m_{ij} \big], \quad m_{ij} = \delta(v_i, v_j), \\
&v_i, v_j \in \{o_1,\dots,o_n, d_1,\dots,d_n\}.
\end{split}
\end{equation}
that accurately reflects real-world traversability by taking into account obstacle constraints,
where $\delta(\cdot,\cdot)$ denotes the obstacle-aware shortest-path distance.

By aligning the clustering phase with the same spatial model used in subsequent path planning (e.g., D*-Lite), our approach ensures consistency across the system. An example of the resulting Dijkstra output is shown in the middle image of Fig.~\ref{fig:spatial_pipeline}. This graph captures obstacle-aware traversal costs and is used in the following clustering and allocation modules in Sec. \ref{subsec:Task assignment}.

\begin{figure}[t]
    \centering
    \includegraphics[width=0.95\columnwidth]{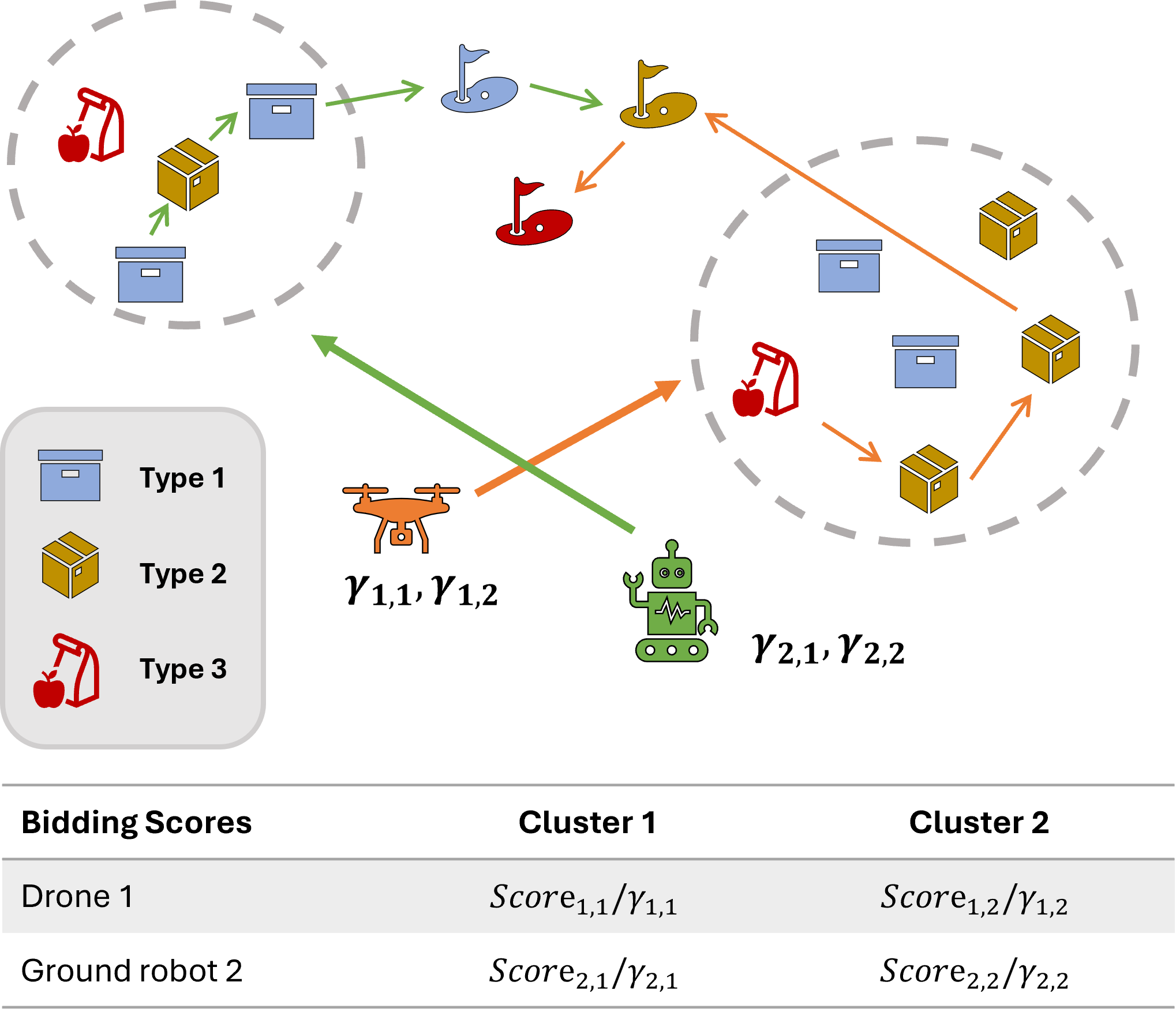}
    \caption{Illustration of the hierarchical task-assignment framework. Tasks are clustered using spatial proximity and obstacle-aware distances. A cluster-level auction assigns task groups to robots based on robot capabilities and the cluster’s task-type distribution, denoted by $\gamma$. Within each assigned cluster, each robot selects tasks subject to capacity limits and task-type compatibility.}
    \label{fig:task_assignment_framework}
\end{figure}

\subsection{Heterogeneous Task Assignment}
\label{subsec:Task assignment}

To address the challenges of task assignment in a heterogeneous robot team with capacity constraints, we propose a novel hierarchical task assignment framework. The overall process operates iteratively as shown by the loop in  Fig.~\ref{fig:oath_framework}: in each round, all currently unassigned tasks are clustered based on the previously computed task-to-task distance matrix; robots then participate in a cluster-level auction, select tasks within their assigned clusters\footnote{Note that, the robot usually will select only a subset of the total tasks within the assigned cluster, due to the limited capacity of the robot.}, and execute them. After completing the current assigned tasks, the robots re-enter the assignment cycle, where the remaining unassigned tasks are clustered again and new intra-cluster task selections are performed. This iterative loop continues until all tasks have been successfully completed. The full procedure consists of three main components: Obstacle-Aware Clustering, Cluster-Weighted Auction, and Intra-Cluster Task Selection. An illustration of this hierarchical task-assignment framework is shown in Fig.~\ref{fig:task_assignment_framework}. The following sections describe each stage in detail.

\begin{algorithm}[t]
\caption{Cluster statistics \& robot--cluster scoring}
\label{alg:cluster_scoring}
\KwIn{Unassigned task set $\mathcal{P}_u$; task-to-task distance matrix $\mathcal{M}$; robot set $\mathcal{R}$; number of task types $|\mathcal{T}|$; robot capability vectors $\{\boldsymbol U_r\in\mathbb{R}_{\ge0}^{|\mathcal{T}|}\}$}
\KwOut{Clusters $\{\mathcal{C}_1,\dots,\mathcal{C}_k\}$; robot-specialization scores $\{\gamma_{r,k}\}$; score matrix $S=\{s_{r,k}\}$}

\textbf{Clustering:}
Cluster all tasks $p \in \mathcal{P}_u$ using agglomerative clustering on $\mathcal{M}$ into clusters $\{\mathcal{C}_1,\dots,\mathcal{C}_k\}$\;

\textbf{Cluster statistics:}
\ForEach{cluster $\mathcal{C}_k$}{
    Compute type-count vector $\boldsymbol N_k\in\mathbb{R}_{\ge0}^{|\mathcal{T}|}$\;
    Define cluster composition: $\psi_k \gets \dfrac{\boldsymbol N_k}{\|\boldsymbol N_k\|_1}$\;
  }

\textbf{Robot preferences:}
  \ForEach{robot $r \in \mathcal{R}$}{
    Normalize capability: $\zeta_r \gets \dfrac{\boldsymbol U_r}{\|\boldsymbol U_r\|_1}$\;
  }

\textbf{Scoring:}
\ForEach{robot $r \in \mathcal{R}$}{
  \ForEach{cluster $\mathcal{C}_k$}{
    Compute distance $\delta_{r,k}$ from $r$ to the center of $\mathcal{C}_k$\;

      Compute robot-specialization score: $\gamma_{r,k} \gets \langle \psi_k,\, \zeta_r\rangle\in(0,1]$\;
      Set score: $s_{r,k} \gets \dfrac{\delta_{r,k}}{\gamma_{r,k}}$\;
    }
  }

\Return{$\{\mathcal{C}_k\}$, $\{\gamma_{r,k}\}$, and $S=\{s_{r,k}\}$}
\end{algorithm}

\begin{algorithm}[t]
\caption{Cluster-Weighted Auction (Assignment Stage)}
\label{alg:cluster_auction}
\KwIn{Robot set $\mathcal{R}$; clusters $\{\mathcal{C}_1,\dots,\mathcal{C}_K\}$; score matrix $S=\{s_{r,k}\}$}
\KwOut{Cluster assignment set $\mathcal{A}$}

Initialize best-bid record $B[k]\gets \infty$ for all $k$, and $\mathcal{A}\gets \emptyset$\;

\Repeat{all robots have conflict-free cluster assignments}{
    \ForEach{robot $r\in \mathcal{R}$}{
        Select cluster index $k^\star \gets \arg\min_k \{\, s_{r,k}\ \mid\ s_{r,k} < B[k]\ \text{and}\ \mathcal{C}_k\ \text{available}\,\}$\;
        \If{$k^\star$ exists}{
            $B[k^\star] \gets s_{rk^\star}$; assign $r \rightarrow \mathcal{C}_{k^\star}$\;
            Resolve conflicts if any and update $\mathcal{A}$\;
        }
    }
}
\Return{$\mathcal{A}$}
\end{algorithm}

\subsubsection{Obstacle-Aware Clustering}

Given the task-to-task distance matrix from Sec. \ref{sec:dijkstra_matric}, 
we group tasks using agglomerative clustering \cite{mullner2011modern}. 
In Algorithm~\ref{alg:cluster_scoring}, Lines~1-4 correspond to this clustering process. 
Unassigned tasks are divided into clusters $\{\mathcal{C}_1, \dots, \mathcal{C}_K\}$ based on the task-to-task distance matrix $\mathcal{M}$.

In general, the environment may contain multiple task types $\tau_i$.
Each cluster $\mathcal{C}_k$ can thus be characterized by the distribution of task types it contains. 
Let $\boldsymbol N_k \in \mathbb{R}_{\ge 0}^{|\mathcal{T}|}$ denote the vector of task counts of each type within $\mathcal{C}_k$. 
To describe the normalized composition of task types in the cluster, we define
\begin{equation}
    \psi_k \;=\; \frac{\boldsymbol N_k}{\|\boldsymbol N_k\|_1}
\end{equation}
To prevent division by zero during the subsequent scoring calculation, a small constant $\theta$ is added to $\boldsymbol N_k$ in the equation above during our implementation. This constant is sufficiently small to minimize its effect on the task assignment results.

This normalized vector $\psi_k$ captures the relative proportion of each task type within the cluster and will be leveraged in the subsequent auction step for scoring. In the next stage, $\psi_k$ is inner-producted with each robot’s capability vector to compute a robot–cluster specialization score that reflects how well a robot matches the task composition of $\mathcal{C}_k$.

\begin{remark}[Obstacle-Aware Distance and Clustering Choice]
    The agglomerative clustering is adopted because it supports arbitrary, non-Euclidean distance metrics, making it suitable for obstacle-rich environments where straight-line measures are often misleading. For instance, $k$-means clustering relies on Euclidean or Manhattan distances, which treat two tasks separated by a wall as “close,” even though the feasible traversable path in reality is actually long. By using obstacle-aware distances computed via Dijkstra on the adaptive Halton map, our method avoids such idealistic groupings. 
\end{remark}

\begin{figure}[t]
    \centering
    \includegraphics[width=0.85\linewidth]{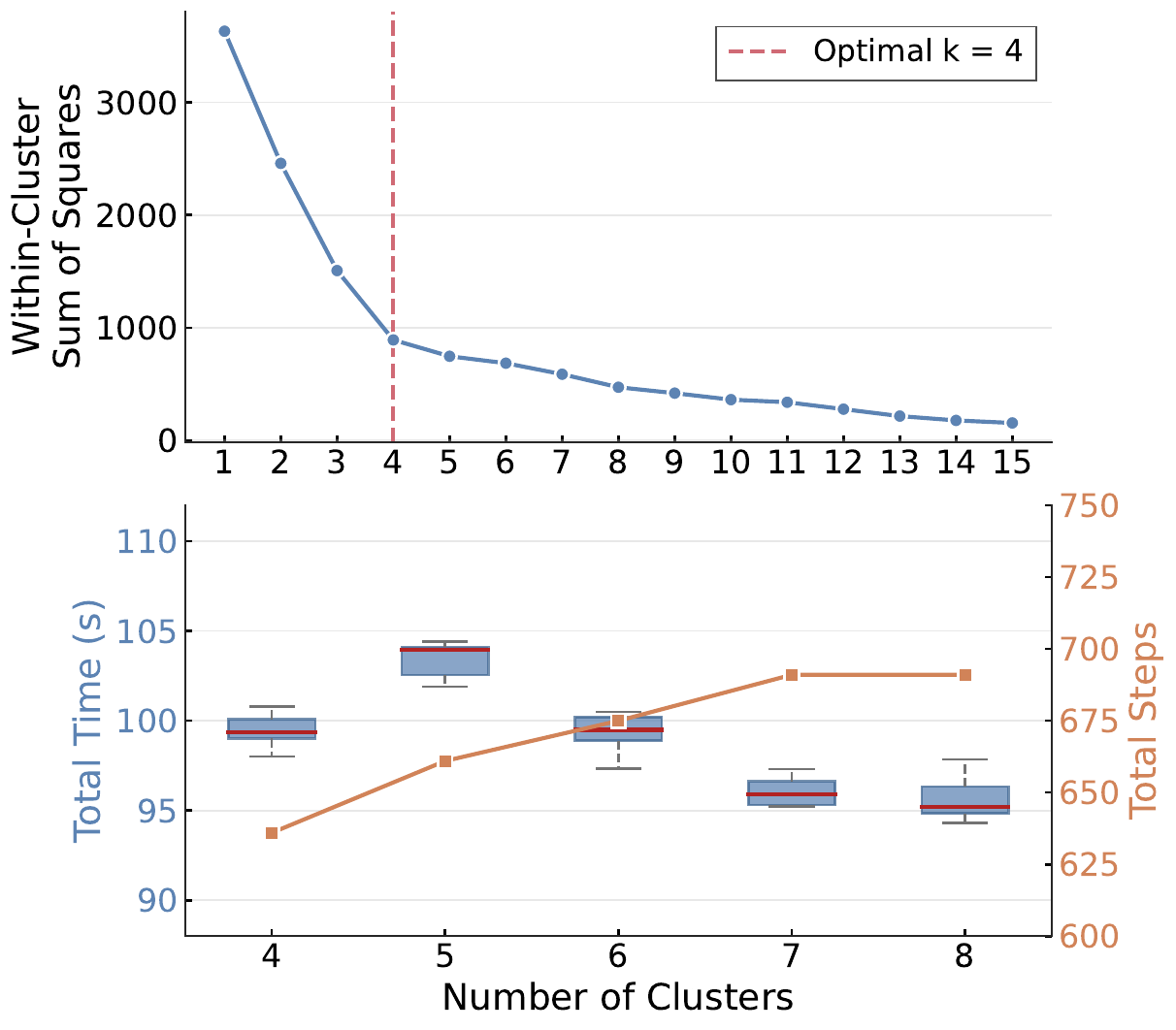}
    \caption{\nan{\textbf{Top:} Elbow-style analysis using the within-cluster sum of squares (WCSS), indicating \(K=4\) as a reasonable choice.
    \textbf{Bottom:} Total completion time (boxplots) and total number of steps (line) under different cluster numbers, showing limited performance variation within a reasonable range of \(K\).}}
    \label{fig:cluster_number}
\end{figure}

\begin{remark}[Selection and Sensitivity of Cluster Number]
    \nan{The number of clusters can be determined using an elbow-style analysis based on within-cluster dispersion, as illustrated in the upper subplot of Fig.~\ref{fig:cluster_number}. 
    In this setting, the within-cluster sum of squares (WCSS) curve exhibits a clear elbow at \(K = 4\), indicating that further increasing the number of clusters leads to diminishing reductions in within-cluster dispersion. 
    Therefore, \(K = 4\) provides an effective trade-off between cluster compactness and task allocation overhead, and is selected as the default cluster number for the experiments conducted in the same map shown in Fig.~\ref{fig:isaac_env}.
    Specifically, we define the sensitivity of a performance metric \(f(K)\) as
    \[
    \mathrm{Sensitivity}(f) =
    \frac{\max_K f(K) - \min_K f(K)}
    {\frac{1}{|\mathcal{K}|}\sum_{K \in \mathcal{K}} f(K)},
    \]
    where \(\mathcal{K} = \{4,5,6,7,8\}\).
    Based on this definition, the sensitivity of the total completion time is approximately \(7.8\%\), and the sensitivity of the total number of steps is approximately \(8.2\%\). 
    These results indicate that OATH is not highly sensitive to the choice of the cluster number within a reasonable range.
    In practice, after selecting an appropriate cluster number, we additionally ensure that the number of clusters is slightly larger than (or at least equal to) the number of robots. 
    This design increases the flexibility of the cluster--auction process and promotes more balanced and localized task groupings across robots. 
    Accordingly, the algorithm enforces that the number of clusters is greater than the number of robots, except when the total number of tasks is smaller than the number of robots. 
    In that case, the number of clusters is set equal to the number of tasks to ensure full task coverage without redundant clustering.}
\end{remark}


\begin{figure*}[t]
    \centering
    \includegraphics[width=\textwidth]{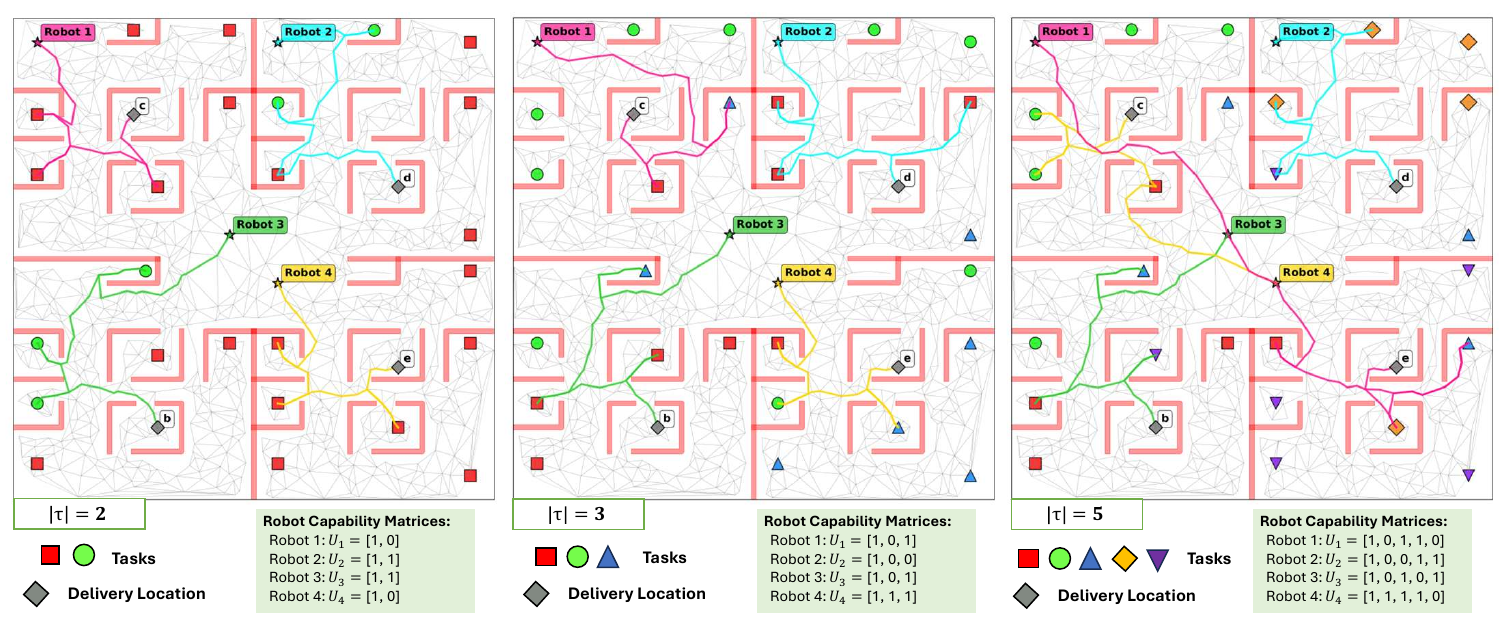}
    \caption{Comparison of heterogeneous task assignment results under different numbers of task types. Each subfigure illustrates the final task assignment for teams operating with $|\mathcal{T}|=2,3,5$, respectively. The results show that each robot strictly follows its capability matrix when selecting tasks. Even when a nearby cluster contains incompatible task types, the robot does not choose it but instead travels to a more distant cluster whose tasks match its capabilities (e.g., in the $|\mathcal{T}|=5$ case, Robot 1 selects a farther cluster that aligns with its feasible task set).}
    \label{fig:heter_task_allo}
\end{figure*}

\subsubsection{Cluster-Weighted Auction (CWA)}

The CWA assigns clusters of tasks to robots by jointly considering spatial proximity and task-type compatibility. 
After task clustering, CWA assigns each robot to a cluster where it is both geographically close (in obstacle-aware distance) and well aligned with the cluster’s task composition.
This process consists of two stages: (i) score computation and (ii) the auction phase.

For score computation phase, each robot $r \in \mathcal{R}$ is represented by a capability vector $\boldsymbol U_r \in \mathbb{R}_{\ge 0}^{|\mathcal{T}|}$, 
We define the normalized preference of each robot as
\begin{equation}
    \zeta_r = \frac{\boldsymbol U_r}{\|\boldsymbol U_r\|_1}.
\end{equation}
For each cluster $\mathcal{C}_k$ with its task-type composition vector $\psi_k$, 
the compatibility between robot $r$ and cluster $\mathcal{C}_k$ is quantified by a robot–cluster specialization factor:
\begin{equation}
    \gamma_{r,k} = \langle \psi_k,\, \zeta_r \rangle \in (0,1],
\end{equation}
which measures how well the robot’s capability distribution aligns with the type composition of the cluster. Here, $\langle \cdot, \cdot \rangle$ denotes the inner product between the two normalized vectors $\psi_k$ and $\zeta_r$.

Each robot then estimates its distance $\delta_{r,k}$ to the cluster center, and combines it with $\gamma_{r,k}$ to form the bidding score:
\begin{equation}
    \text{score}_{r,k} = \frac{\delta_{r,k}}{\gamma_{r,k}},
\end{equation}
such that robots closer to the cluster and better matched to its task distribution have a higher probability to win the auction.  Algorithm~\ref{alg:cluster_scoring} (Lines~5-12) implements this scoring step.

Once all scores are calculated, the auction proceeds in a sequential manner. Robots are ordered (e.g., by robot index number), and each robot selects the cluster with the best (lowest) score that is still available. If a robot selects a cluster that has already been claimed by another robot, the previous assignment is revoked, and that robot will re-enter the auction in the next iteration. Algorithm~\ref{alg:cluster_auction} implements this auction step.

This bidding and conflict resolution process continues until all robots are assigned to non-overlapping clusters. The final assignment result is stored in the cluster assignment set $\mathcal{A}$, which maps each robot to a task cluster for subsequent intra-cluster planning.

\subsubsection{Intra-Cluster Task Selection}

After tasks have been grouped into smaller clusters and assigned to individual robots, each robot selects a subset of type-compatible tasks from its assigned cluster, within its capacity limit. The goal is to determine the locally optimal sequence of pickup and delivery locations that minimizes travel cost. This process can be formulated as a small-scale Traveling Salesman Problem (TSP) with constraints, which is efficiently solvable as MILP using an off-the-shelf optimization solver (e.g., Gurobi). The formal problem definition is given below.

\vspace{0.5em}
\noindent
\textbf{Problem 2.}
Given a robot $r$ with capacity $Q_r$ and an assigned task set $\mathcal{C}_r = \{1,2,\dots,n_r\}$, where $n_r$ is the number of tasks in the cluster, the objective is to determine the optimal pickup–delivery sequence over obstacle-aware locations that minimizes total travel cost, subject to task type compatibility, capacity, and precedence constraints.

Let $\mathcal{V}_r = \{v_1, v_2, \dots, v_{n_r}, \dots, v_{n_r+m_r}\}$ denote the set of all pickup and delivery locations associated with tasks in $\mathcal{C}_r$, where $n_r$ is the number of pickups (equal to the number of tasks) and $m_r$ is the number of deliveries. 
Define $x_{ij} \in \{0,1\}$ as a binary variable indicating whether the robot travels directly from $v_i$ to $v_j$, $\ell_i$ as the carried load upon arrival at $v_i$, and $a_i$ as the arrival time at $v_i$. $u_i$ is an integer ordering variable associated with location $v_i$, 
indicating its position in the tour (i.e., the visit order); and $\delta(i,j)$ is the obstacle-aware travel cost from location $v_i$ to $v_j$, defined previously in Sec. \ref{sec:dijkstra_matric}.

Intra-cluster task selection for robot $r$ is formulated using MILP:

\begin{subequations}\label{eq:opt_problem}
\begin{alignat}{2}
\min_{x} \; & \sum_{v_i \in \mathcal{V}_r} \sum_{v_j \in \mathcal{V}_r} \delta(i,j)\, x_{ij} \label{eq:obj} \\
\text{s.t.}\; 
& \sum_{v_j \in \mathcal{V}_r} x_{ij} = 1,  &&\forall v_i \in \mathcal{V}_r \label{eq:flow_out} \\
& \sum_{v_i \in \mathcal{V}_r} x_{ij} = 1, &&\forall v_j \in \mathcal{V}_r \label{eq:flow_in} \\
& 0 \leq \ell_i \leq Q_r, &&\forall v_i \in \mathcal{V}_r \label{eq:capacity} \\
& a_{p_i} < a_{d_i}, &&\forall v_i \in \mathcal{V}_r \label{eq:pickup_before_delivery} \\
& x_{ij} = 0, &&\text{if }(\boldsymbol U_r)_{\tau_{i/j}}=0
\label{eq:compatibility} \\
& u_i + 1 \leq u_j + |\mathcal{V}_r| \, (1 - x_{ij}), &&\forall i \neq j,\; v_i,v_j \in \mathcal{V}_r
\label{eq:subtour}
\end{alignat}
\end{subequations}
Constraints~\eqref{eq:flow_out} and~\eqref{eq:flow_in} ensure that each location is visited exactly once. 
Constraint~\eqref{eq:capacity} imposes the robot’s capacity limit $Q_r$; 
Constraint~\eqref{eq:pickup_before_delivery} guarantees that each pickup precedes its corresponding delivery; 
Constraint~\eqref{eq:compatibility} eliminates infeasible assignments based on robot–task compatibility; and 
Constraint~\eqref{eq:subtour} eliminates subtours and ensures route connectivity. 
We adopt the Miller--Tucker--Zemlin (MTZ) formulation~\cite{miller1960integer}, 
a classical linearization technique widely used in TSP/VRP problems. 
In the MTZ formulation, the auxiliary variables $u_i$ represent the visiting order of each location, same as we do. 
Specifically, Constraint~\eqref{eq:subtour} enforces that if edge $(i,j)$ is included in the route ($x_{ij}=1$), 
then node $j$ must appear after node $i$ in the visiting sequence; 
when $x_{ij}=0$, the inequality is relaxed by the large constant term $|\mathcal{V}_r|$. 
Together, this constraint prevents the formation of disconnected loops 
and guarantees a single continuous tour for each robot.

The results of the heterogeneous, obstacle-aware task assignment are illustrated in Fig.~\ref{fig:heter_task_allo}. The figure presents three allocation outcomes under different numbers of task types, $|\mathcal{T}| = 2, 3,$ and $5$, respectively. Each robot is assigned to a cluster of tasks that are both spatially accessible in free space (i.e., regions without long wall segments) and compatible with the robot’s capability vector. In particular, when the nearest cluster contains incompatible task types, the robot selects a more distant but feasible cluster instead, as observed in the $|\mathcal{T}| = 5$ case where Robot~1 chooses a further cluster that aligns with its task-handling abilities. These results demonstrate that the proposed heterogeneous task assignment algorithm effectively balances spatial proximity and capability matching, enabling each robot to select reachable and compatible task clusters while maintaining obstacle-aware and efficient allocations.

\begin{remark}[Global Optimality and Replanning Trade-off]
Within each cluster, the MILP produces an optimal sequence for the tasks assigned to a single robot under the current cluster composition and capacity constraints. The overall framework, however, does not guarantee global optimality: cluster assignments and routes are determined iteratively and may become suboptimal as execution unfolds. Increasing the frequency of the cluster–auction–intra-cluster selection cycle can improve realized path quality by re-optimizing more often, but it also incurs additional task reallocation and replanning overhead, which may increase wall-clock mission time. In practice, there is a trade-off between solution quality and computational efficiency.
\end{remark}

\begin{remark}[Motion Heterogeneity]
\nan{Beyond the binary capability vector $U_r$, OATH supports motion heterogeneity through robot-specific transition systems. During roadmap construction, we define robot-specific feasibility constraints and edge costs. Differences in obstacle-surmounting ability are encoded either by adjusting edge weights or by removing infeasible transitions. Since allocation relies on robot-specific shortest-path distances, these differences naturally propagate to cluster formation and route selection without modifying the core framework.}
\end{remark}

\begin{figure}[t]
    \centering
    \includegraphics[width=0.8\linewidth]{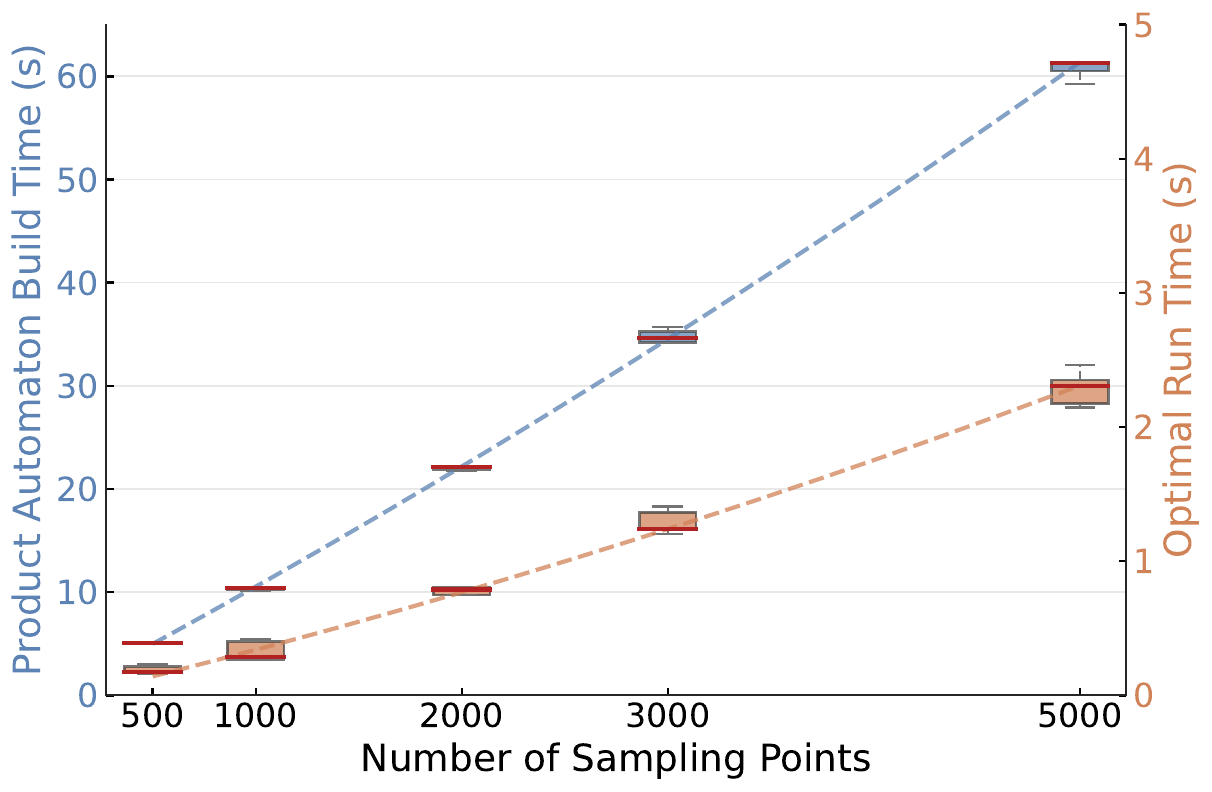}
    \caption{\nan{Sensitivity analysis of product automaton construction time and optimal-run search time with respect to the number of sampling points. 
    Blue boxplots represent the product automaton construction time, and orange boxplots represent the optimal-run computation time. 
    For each sampling density, the experiment is repeated five times using the same LTL specification but different random seeds, resulting in distinct sampling configurations. Results are summarized using boxplots.}
    }
    \label{fig:automaton time}
\end{figure}

\begin{figure*}[t]
    \centering
    \includegraphics[width=\textwidth]{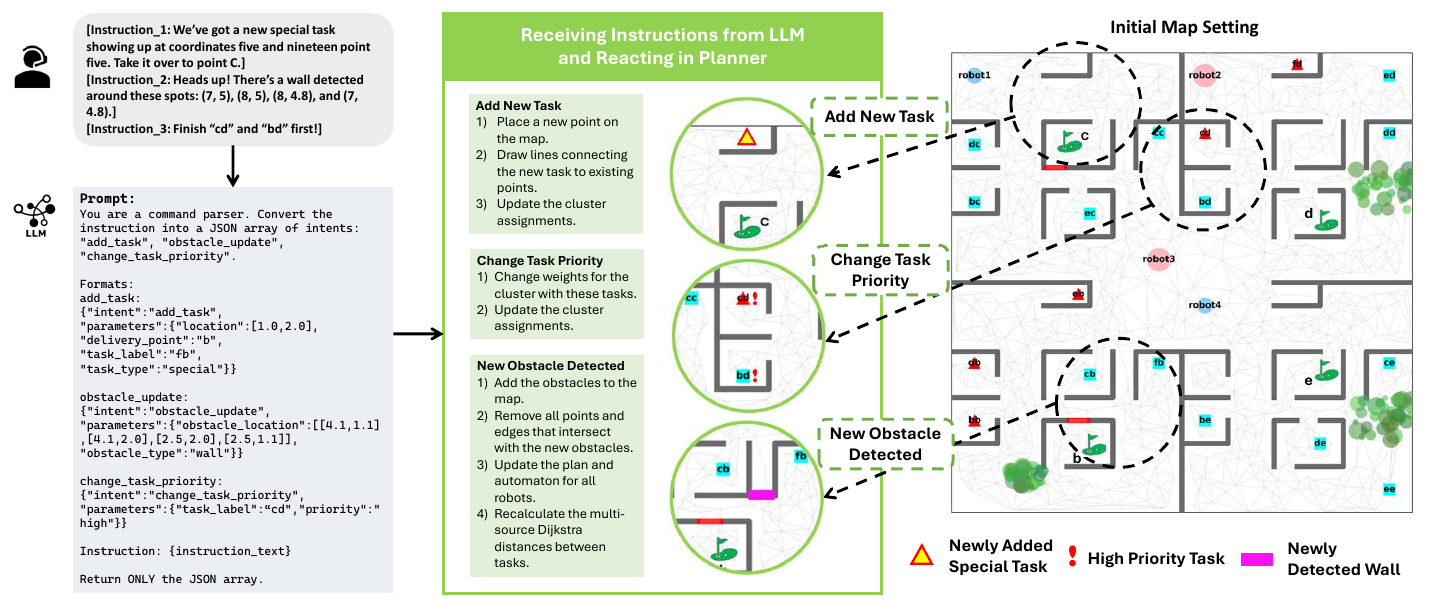}
    \caption{Overview of how the LLM parses human instructions and how the planner reacts. 
    The LLM converts natural language into structured intents (\texttt{add\_task}, \texttt{obstacle\_update}, \texttt{change\_task\_priority}). 
    The planner then updates the map and task assignment accordingly (center panel). 
    The right panel shows an example environment with $4$ robots (two drones and two ground robots), delivery tasks (blue squares), inspection tasks (red triangles), delivery points (green flags), and bushes (green circles). 
    Letter pairs shown next to tasks (e.g., \texttt{dc}, \texttt{ec}, \texttt{bb}) are task labels used by the planner and in instructions.}
    \label{fig:llm_planner_overview}
\end{figure*}

\subsection{Dynamic Path Planning}

Once each agent has received its individual task assignment, it must execute the assigned tasks in a specific order while avoiding unintended goal locations. To enforce temporal constraints and ensure correctness of task execution, we employ Linear Temporal Logic (LTL) to formally encode the required task sequences for each agent. LTL provides a compact language for specifying ordering and safety requirements, such as ``task $A$ must eventually follow task $B$'' or ``task $C$ must never be visited before task $D$.''

For each agent, we automatically construct an LTL formula based on its assigned tasks. This formula is then translated into a Büchi automaton \cite{gastin2001fast}, a finite-state machine that accepts infinite sequences satisfying the LTL specification. The Büchi automaton is then composed with a transition system that models the environment, yielding a product automaton. The product automaton captures both the temporal requirements and the spatial structure of the workspace, ensuring that any path found is correct with respect to the assigned specification.

Path planning is then carried out on the product automaton using the D*-Lite algorithm~\cite{ren2024ltl,koenig2005fast}, a dynamic and heuristic search method that supports efficient replanning. This allows agents to update their paths during execution in response to newly detected or previously unknown obstacles, maintaining the execution feasibility while adapting to changes in the environment.

\begin{figure*}[t]
    \centering
    \includegraphics[width=\textwidth]{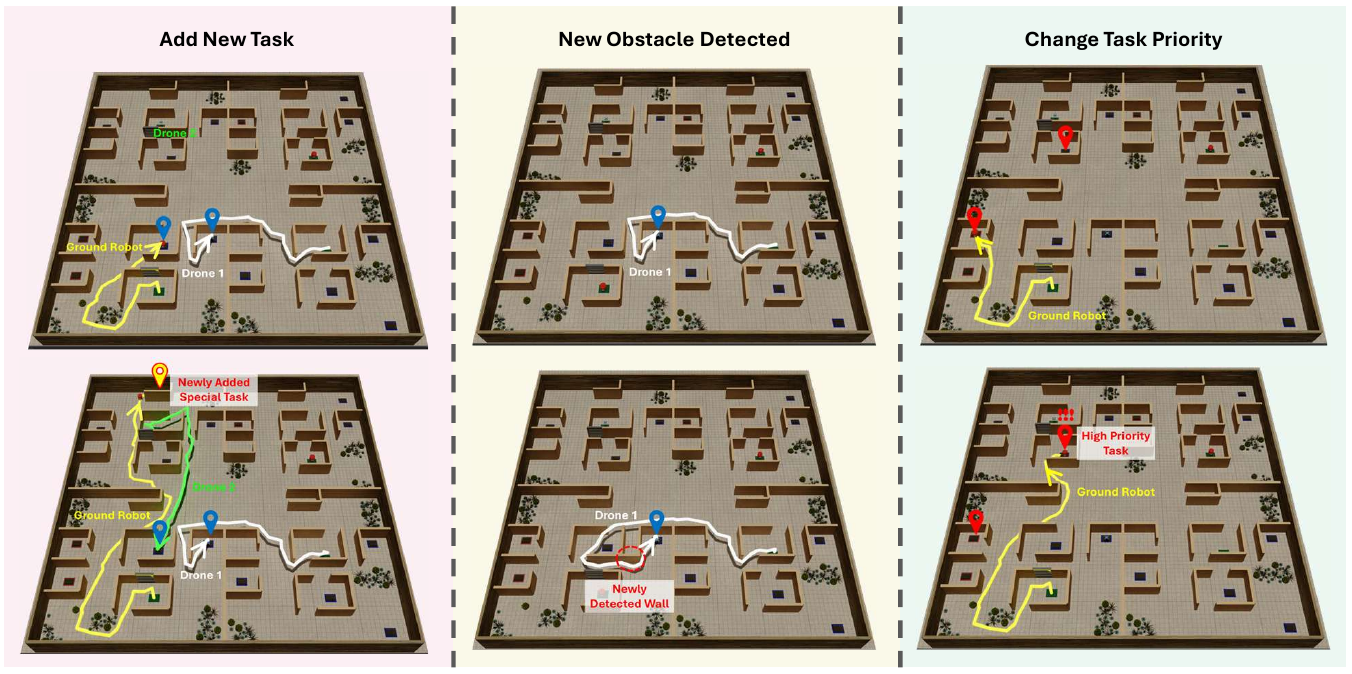}
    \caption{Demonstration of obstacle-aware task assignment and route replanning under LLM-parsed human instructions in Isaac Sim. The first row shows the initially planned routes, while the second row shows the updated trajectories after executing new instructions: (left) adding a new task, (middle) detecting a new obstacle, and (right) changing task priority. The adaptive planner dynamically adjusts robot paths and task assignments in response to each instruction.}
    \label{fig:LLM_route}
\end{figure*}

\begin{remark}[Product Automaton Construction Cost]
The time required to construct the product automaton depends on two main factors. 
First, the complexity of the LTL formula increases with the number of tasks encoded; for example, planning a route for ten tasks results in a significantly longer specification than planning for only three.
Second, the size of the environment model which is determined by the number of sampled points in the underlying map, directly affects the number of states in the transition system and thus the computational cost of constructing the product automaton. In our case, sampling $1000$ adaptive Halton points generates a transition system that is combined with a Büchi automaton containing $5$ states and $12$ transitions. The resulting product automaton used for path planning has about $20,520$ states and $294,792$ transitions. If the LTL specification remains the same but the number of sampled points increases to $2000$, the resulting product automaton expands to about $40,520$ states and $613,800$ transitions, causing the construction time to roughly double. \nan{To further illustrate the time consumption of product automaton construction and optimal-run search under increasing sampling densities, Fig.~\ref{fig:automaton time} reports a sensitivity analysis.
The blue boxplots shows the time required to build the product automaton, while the orange boxplots shows the time required to find an optimal run.
As the number of sampling points increases, both components exhibit a clear increasing trend, with the construction of the product automaton dominating the overall runtime.}
Since each agent must rebuild its product automaton after every task assignment cycle, this construction cost is a critical factor. 
Our use of the adaptive Halton sequence provides an additional advantage: in cluttered regions, higher sampling density captures obstacle geometry sufficiently to guarantee feasible paths, while in open areas, points are sampled at lower density and in fewer numbers. 
By reducing unnecessary points in open regions, the adaptive map helps control the size of the product automaton and thus shortens automaton synthesis time.
\nan{Although incremental construction of the product automaton is desirable in general, it is not directly applicable in our task allocation setting.
In OATH, task allocation and path planning are performed in an iterative closed loop, where each allocation round produces a new LTL specification for each robot based on its updated task sequence.
As a result, the Büchi automaton itself changes across allocation cycles, which precludes reuse or incremental update of a previously constructed product automaton.
Nevertheless, incremental updates are exploited whenever the task-level specification remains unchanged.
In particular, during execution-level replanning triggered by environmental changes (e.g., newly detected obstacles), the Büchi automaton is fixed, and replanning is performed by locally modifying the product automaton through edge invalidation and incremental cost repair using D*-Lite.
This design ensures that full Büchi–product automaton reconstruction is performed only when task-level temporal specifications change, while incremental updates are used whenever possible to reduce replanning overhead.}
\end{remark}


\section{Large Language Model as Translator}
\label{sec:llm_translator}

Different from previous methods that only use large language models (LLMs) as an offline planner or as a translator in the pre-execution stage, our framework integrates the LLM as a persistent interpreter throughout the whole task execution process. In this way, the LLM becomes an always-available interface between human instructions and the planning system.

As shown in Fig.~\ref{fig:llm_planner_overview}, the interaction between the LLM and the planner follows a clear pipeline. After receiving a new instruction, the LLM first determines the intent (e.g., adding a task, updating an obstacle or changing task priority) and extracts key parameters such as positions, labels, or constraints. These parameters are converted into a JSON structure and passed to the high-level planner. The planner then updates the map and the task graph according to a predefined pipeline.

In detail, the instruction “We have got a new special task showing up at coordinates five and nineteen point five. Take it over to point C.” is recognized as an \textit{Add New Task} command. The LLM extracts the task’s pickup coordinates $(5, 19.5)$, the delivery point $C$, and auxiliary fields such as task type. Rather than rebuilding the entire map, the planner performs an incremental update: it keeps the existing adaptive Halton sequence roadmap, inserts a new task node at $(5,19.5)$, and uses Delaunay triangulation (i.e., connecting the node to its nearest spatial neighbors while preserving non-overlapping edges) to maintain geometric consistency. The new task node is then incorporated in the next cluster–auction–intra-cluster cycle; when robots rebuild their product automata, they automatically integrate the new coordinates, while robots currently executing tasks remain unaffected. This update process typically completes within about five seconds in our simulation setup. \nan{This 5-second delay refers to the end-to-end system response time rather than the LLM inference time alone. The total response time includes roadmap update, distance matrix computation, clustering, task reassignment, and path replanning.}

For a constraint-oriented instruction such as “Heads up! There’s a wall detected around these spots: $(7, 5)$, $(8, 5)$, $(8, 4.8)$, and $(7, 4.8)$,” the LLM interprets it as an \textit{New Obstacle Detected} command. The detected obstacle coordinates are passed to the planner, which removes affected nodes and edges, updates the map, and replans the route to avoid the new wall. 

Similarly, for an instruction \textit{Change Task Priority}, the planner adjusts the weight of the affected cluster and updates the corresponding task assignments. 
The planner’s response and its impact on robot routes are illustrated in Fig.~\ref{fig:LLM_route}. The top row shows the original task assignments and planned routes, while the bottom row depicts the updated trajectories after executing the new LLM-parsed instructions. In the \textit{Add New Task} scenario, the ground robot is initially assigned to a nearby task. When the new special task appears, which can only be handled by the ground robot, the system immediately reassigns it to complete this task first, even though it is farther away. 
In the \textit{New Obstacle Detected} case, the drone dynamically adjusts its trajectory to circumvent the obstacle without redundant exploration.
In the \textit{Change Task Priority} scenario, when a distant task is assigned a higher priority, the ground robot modifies its plan and proceeds to complete that task first, reflecting an immediate reordering of objectives.

Overall, these examples demonstrate that the LLM-guided planner can seamlessly integrate new high-level instructions into ongoing task assignment and path planning. Human operators can intervene at any time without interrupting the execution process. This allows language-driven updates to tasks, environments, and constraints to be processed quickly and flexibly, which is crucial for planning in dynamic and uncertain environments.

\nan{To support more natural and less structured user inputs, we design an interactive human--LLM command interface.
Rather than requiring fully specified commands, the interface allows users to issue high-level or ambiguous instructions and engage in a short clarification dialogue with the LLM.
The LLM parses the intent of each instruction, identifies any parameters required by the planner that are missing or ambiguous, and explicitly requests additional information from the user when needed.
The interface illustration and the complete prompt details are provided in Appendix~\ref{app:llm_ui}.}

\section{Experiments}\label{sec:experiments}

We evaluate OATH on obstacle-rich maps (Fig.~\ref{fig:isaac_env}) containing both known walls and unknown obstacles (e.g., iron gates and bushes). 
An adaptive Halton sequence is generated to capture the map structure and provide sampling points. 
We compare our method OATH against three alternatives, including two ablation experiments and a CBBA\cite{choi2009consensus} baseline. 
All methods use the same map instantiation, robot team, and task set, and rely on LTL-D*\cite{ren2024ltl} for path planning, so differences arise solely from task assignment algorithm. 
We verify the proposed framework in NVIDIA Isaac Sim~\cite{nvidia_isaac_sim}, demonstrating that the approach can be integrated into a realistic simulation environment.

\subsection{Methods Compared}

\begin{enumerate}
  \item \textbf{CBBA}: the consensus-based bundle algorithm for task assignment.
  \item \textbf{K-means+Auction+NN (KAN)}: $k$-means clustering, cluster-level auction, and a nearest-neighbor (NN) heuristic for intra-cluster task ordering with Euclidean distances.
  \item \textbf{K-means+Auction+MILP (KAM)}: $k$-means clustering, cluster-level auction, with an MILP-based intra-cluster solver with Dijkstra distances.
  \item \textbf{OATH (ours)}: obstacle-aware clustering on the adaptive Halton map, cluster-weighted auction (CWA), and MILP-based intra-cluster selection.
\end{enumerate}
\begin{table}[t]
\centering
\caption{Comparison of methods evaluated. All variants use the same path planning algorithm.}
\label{tab:methods}
\begin{tabular}{@{}lccc@{}}
\toprule
\textbf{Method} & \textbf{Clustering} & \textbf{Intra-Cluster} \\ \midrule
CBBA & Bundle & Bundle Order \\
KAN & $k$-means & NN \\
KAM & $k$-means & MILP \\
OATH (ours) & Obstacles-Aware Cluster & MILP \\ \bottomrule
\end{tabular}
\end{table}

\subsection{Metrics}\label{subsec:metrics}

We evaluate performance using three metrics and multiple independent randomized task placement trials with identical random seeds across all methods.

\begin{enumerate}[label=\arabic*)]
\item \textbf{Task Assignment Time} ($T_{\text{alloc}}$): wall–clock time spent by the allocator \emph{before} execution begins. 
This metric focuses purely on the computational efficiency of the allocation algorithm and indicates how quickly each method can produce an allocation in real time, independent of the resulting solution quality.

\item \textbf{Total Running Steps} ($S_{\text{total}}$): the sum of discrete execution steps taken by all robots until all tasks are completed,
\[
S_{\text{total}}=\sum_{i=1}^{N} s_i,
\]
where $s_r$ is the number of steps executed by robot $r$. Since all methods use the same low-level planner (LTL-D* \cite{ren2024ltl}), $S_{\text{total}}$ primarily reflects the efficiency of task assignment and intra-cluster selection, where more optimal allocators enable the team to complete all tasks with fewer overall movements.

\item \textbf{Total Running Time} ($T_{\text{total}}$): end-to-end wall–clock mission time from the start of allocation to the completion of the last task,
\[
T_{\text{total}} = T_{\text{alloc}} + T_{\text{plan+exec}},
\]
where $T_{\text{plan+exec}}$ includes the time of LTL-D* planning/replanning and simulated execution. This metric evaluates not only the optimality of the task assignment results but also accounts for the time spent on task assignment, automaton construction, and replanning, and therefore it reflects the overall efficiency of the proposed algorithm.
\end{enumerate}

\begin{figure*}[t]
    \centering
    \includegraphics[width=\textwidth]{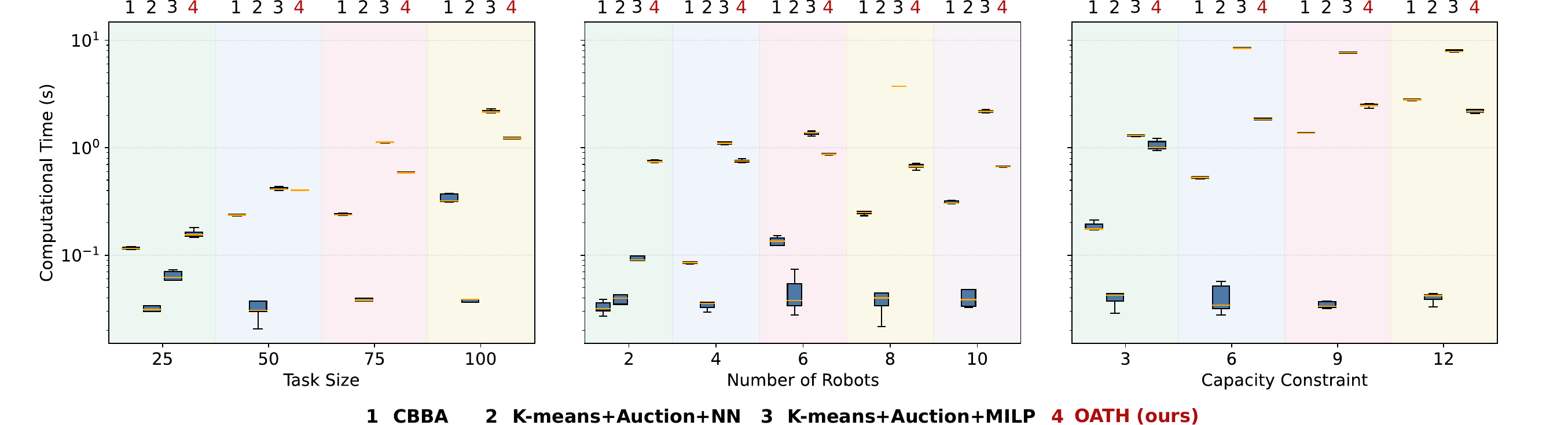}
    \caption{Computational time of the task assignment algorithm under different experimental settings. Left: varying the task size while keeping the number of robots fixed at $10$ and the capacity fixed at $5$. Middle: varying the number of robots while keeping the task size fixed at $100$ and the capacity fixed at $5$. Right: varying the capacity while keeping the task size fixed at $120$ and the number of robots fixed at $10$. All results are shown as boxplots to illustrate the distribution of running times across different scenarios.}
    \label{fig:task_allocation_compare}
\end{figure*}

\subsection{Experimental Design}
We organize our evaluation into four groups of studies, targeting complementary aspects of the system: allocation efficiency, end-to-end performance, scalability, and LLM-guided instruction handling.

\begin{enumerate}[label=\arabic*)]

\item \textbf{Task Assignment Algorithm Efficiency.} We conduct three controlled experiments and test $T_{\text{alloc}}$ for each algorithm:  
\begin{enumerate}[label=\alph*)]
    \item \emph{Varying Task Number:} fix the number of robots $R$ and the per-robot capacity $Q$, and increase the number of tasks $P$ to examine sensitivity to problem size.  
    \item \emph{Varying Robot Number:} fix $P$ and $Q$, and increase the number of robots $R$ to examine sensitivity to team size.  
    \item \emph{Varying Capacity:} fix $R$ and $P$, and vary the per-robot capacity $Q$. Larger $Q$ increases the complexity of assignment, directly testing allocator robustness.  
\end{enumerate}

\item \textbf{End-to-End Performance.}
\nan{
To evaluate the end-to-end performance of our framework, we conduct two sets of experiments.}
\begin{enumerate}[label=(\arabic*)]
    \item We assess the overall system performance, including \emph{map construction, task assignment, path planning, and execution}.
    \item \nan{We assess the solution quality of OATH on small-scale problem instances by comparing the total execution distance against an optimization-based MILP baseline, which is solved directly using a commercial solver.}
\end{enumerate}

\item \textbf{Scalability of OATH.}  
We evaluate the scalability of the proposed method (OATH) under large task regimes through two experiments.
\begin{enumerate}[label=(\arabic*)]
    \item \nan{We examine the MILP solver time with respect to robot task capacity.}
    \item We increase the number of tasks \(P\) from $25$ to $100$ on the Halton map and measure both the total number of steps \(S_{\text{total}}\) and the total completion time \(T_{\text{total}}\).
\end{enumerate}

\item \textbf{LLM-Guided Instruction Performance.}  
\nan{We evaluate the performance of the LLM-guided instruction module through two experiments. 
The first experiment measures the intent parsing accuracy of the LLM parser, and the second examines the impact of LLM-guided human instructions on overall system behavior.}

\nan{(i) To evaluate intent parsing accuracy, we construct a test set of $90$ natural-language instructions, consisting of three intent categories: \textit{add new task}, \textit{new obstacle detected}, and \textit{change task priority}, with $30$ instructions per category. For each intent, the instructions include both clearly specified commands and intentionally ambiguous, colloquial commands to reflect realistic human inputs. Examples include a clear instruction such as ``Create a new task at $(5, 16)$ and send it to room B,'' and a more ambiguous one such as ``A new job popped up close to room C.'' Similar clear and ambiguous instructions are designed for obstacle updates and priority changes. All instructions are processed by the LLM parser, and we record the predicted intent for each input.}

(ii) We further investigate the impact of LLM-guided human instructions on system behavior. We consider three representative instruction types:  \emph{Add New Tasks}, \emph{New Obstacles Detected}, and \emph{Change Task Priority}. For each type, we compare two modes: a \emph{system built-in baseline} and an \emph{LLM-guided setting}. All experiments are conducted with $R=4$ robots and $P=25$ tasks. Each condition is repeated five times, with human instructions intervened at varying time points. We report both $S_{\text{total}}$ and $T_{\text{total}}$ as evaluation metrics to capture efficiency and solution quality. The designs are as follows:  
\begin{enumerate}[label=\alph*)] 
    \item \emph{Add New Tasks.} In the baseline, the full set of $25$ tasks is available from the start. In the LLM-guided setting, only $24$ tasks are initialized, and one additional task is introduced during execution via human instruction.
    \item \emph{New Obstacles Detected.} In the baseline, the map is initialized with the obstacles already present. In the LLM-guided setting, the obstacles are initially absent and introduced dynamically during execution. 
    \item \emph{Change Task Priority.} In the baseline, task priorities remain fixed. In the LLM-guided setting, a human instruction dynamically modifies the priority of one task during execution.  
\end{enumerate}

\nan{In addition to evaluating intent parsing accuracy and system-level performance impact, we undertake a user study to quantify human-in-the-loop efficiency and analyze the latency characteristics of the LLM-guided pipeline. The detailed experimental setup and corresponding results are provided in Appendix~\ref{app:user_study}.}

\end{enumerate}

\section{Results}

\subsection{Parameters Used in Experiments}

As shown in Fig.~\ref{fig:spatial_pipeline}, we summarize in Table~\ref{tab:exp_params} the parameter settings used throughout all experiments. 
These values are fixed unless otherwise specified.
\begin{table}[h]
\centering
\caption{Parameters used in experiments}
\label{tab:exp_params}
\begin{tabular}{l c c}
\toprule
\textbf{Parameter} & \textbf{Symbol} & \textbf{Value} \\
\midrule
Minimum safe distance & $\delta_{\min}$ & 0.3 \\
Optimal sampling distance & $\delta_{\text{opt}}$ & 0.4 \\
Gaussian width & $\sigma$ & 0.5 \\
Baseline acceptance probability & $\beta$ & 0.2 \\
Number of Halton points & -- & 2000 / 1000 \\
Type density weight & $\theta$ & 0.1 \\
\bottomrule
\end{tabular}

\vspace{0.5em}
{\footnotesize \textit{Note:} Halton points = 2000 for task assignment and scalability experiments; 1000 for end-to-end and LLM experiments.}

\end{table}

\subsection{Results on Task Assignment Algorithm Efficiency}
In the first group of experiments, we evaluate the computational efficiency of the task assignment algorithms under increasing complexity. Three separate studies are conducted by varying one of the following parameters while keeping the others fixed: $P$, $R$, and $Q$. The results are shown in Fig.~\ref{fig:task_allocation_compare}.

Although OATH does not achieve the lowest absolute runtime in simple cases, its runtime growth is much flatter compared with CBBA and KAM. 
To quantify this observation, we define a \emph{complexity sensitivity index} as
\begin{equation}
    \text{Sensitivity} = \frac{T_{\max} - T_{\min}}{\Delta x},
\end{equation}
where $T_{\max}$ and $T_{\min}$ are the runtimes under the most complex and simplest configurations, respectively, and $\Delta x$ denotes the corresponding change in the complexity parameter (e.g., $P$, $R$, or $Q$). 
A lower value indicates that the method’s runtime grows more gently as the problem complexity increases.

\begin{table}[h]
\centering
\caption{Complexity sensitivity of different allocators, defined as $(T_{\max} - T_{\min}) / \Delta x$ under varying task number, robot number, and capacity. Lower values indicate better robustness to complexity growth.}
\label{tab:sensitivity}
\begin{tabular}{lccc}
\toprule
\textbf{Method} & \textbf{Task Number ($P$)} & \textbf{Robot Number ($R$)} & \textbf{Capacity ($Q$)} \\
\midrule
CBBA                  & \textbf{\textcolor{red}{0.003}} & 0.035 & 0.296 \\
KAN    & \textbf{\textcolor{red}{5.04E-05}}  & \textbf{\textcolor{red}{0.001}} & \textbf{\textcolor{red}{0.001}}  \\
KAM  & 0.028 & 0.261 & 0.797 \\
OATH                  & 0.015 & \textbf{\textcolor{red}{0.010}} & \textbf{\textcolor{red}{0.13}} \\
\bottomrule
\end{tabular}
\end{table}

Table~\ref{tab:sensitivity} summarizes the sensitivity scores. 
KAN attains the smallest sensitivity overall, but it is a very simple pipeline that prioritizes speed over solution quality: the \(k\)-means pre-clustering and nearest-neighbor assignment do not optimize an objective, provide no optimality guarantees, and are prone to cluster-imbalance effects; moreover, the consistency of distance matrices used in task assignment and path planning is largely ignored. 
OATH is not always among the two smallest entries in all matrices; however, its scores remain consistently low — the maximum is only $0.015$ — so it is the second smallest overall. 
Compared with CBBA and KAM, OATH is typically about an order of magnitude less sensitive in most cases, indicating that its allocation module scales more gracefully with problem growth.

\begin{table}[h]
\centering
\caption{Total steps and total running time (s) vs.\ task number.}
\label{tab:steps_time}
\begin{tabular}{lrrrr}
\toprule
\multicolumn{5}{c}{\textbf{Total Steps}}\\
\midrule
Tasks & CBBA & KAN & KAM & OATH \\
\midrule
10 & 314 & 332 & 330 & \textcolor{red}{\textbf{222}}\\
15 & 667 & 451 & 456 & \textcolor{red}{\textbf{351}}\\
20 & 773 & 555 & 510 & \textcolor{red}{\textbf{463}}\\
25 & 954 & 654 & 608 & \textcolor{red}{\textbf{562}}\\
30 & 1031 & 820 & 802 & \textcolor{red}{\textbf{768}}\\
\midrule
\multicolumn{5}{c}{\textbf{Total Running Time (s)}}\\
\midrule
Tasks & CBBA & KAN & KAM & OATH \\
\midrule
10 & 68.53 & 54.79 & 54.60 & \textcolor{red}{\textbf{40.36}}\\
15 & 94.20 & 84.13 & 83.30 & \textcolor{red}{\textbf{60.11}}\\
20 & 106.02 & 90.66 & 85.96 & \textcolor{red}{\textbf{76.98}}\\
25 & 111.96 & 95.02 & 95.35 & \textcolor{red}{\textbf{94.76}}\\
30 & 124.94 & 113.14 & 129.95 & \textcolor{red}{\textbf{103.63}}\\
\bottomrule
\end{tabular}
\end{table}

\subsection{Results on End-to-End Computational Performance}

The second group of experiments evaluates the overall planner performance, including map construction, task assignment, planning, and execution. Two metrics are considered: $T_{\text{total}}$ and $S_{\text{total}}$.

\begin{figure}[t]
    \centering
    \includegraphics[width=0.8\linewidth]{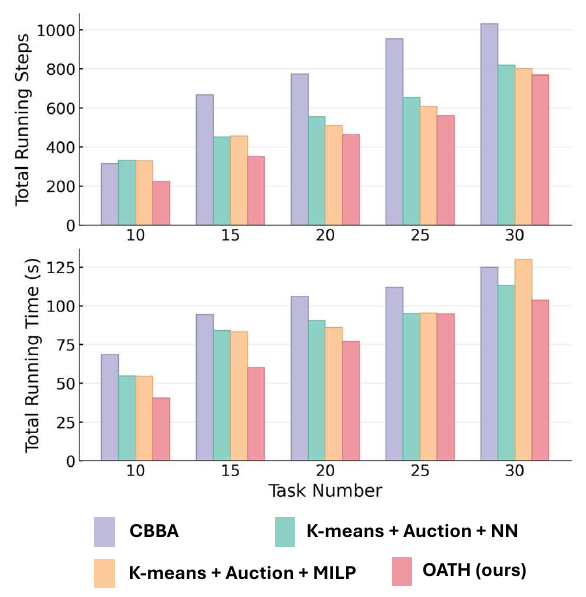}
    \caption{Experimental results of the task assignment algorithm in a simulated environment.  Top: total running steps of the four robots as a function of task number. Bottom: total running time of the four robots as a function of task number.}
    \label{fig:total_running}
\end{figure}

\begin{table*}[t]
\centering
\renewcommand{\arraystretch}{1.3}
\caption{\nan{Comparison between OATH and the MILP baseline solved using Gurobi in terms of total execution distance and computation time.}}
\label{tab:steps_time_gap}
\begin{tabular}{c|cccc|cc}
\hline
\multirow{2}{*}{Number of Tasks} 
& \multicolumn{4}{c|}{Total Distance}
& \multicolumn{2}{c}{Time (s)} \\
\cline{2-5} \cline{6-7}
& OATH 
& MILP 
& Gap to MILP (\%) 
& Gap to LB (\%)
& OATH 
& MILP \\
\hline
9  & 88.108 & 67.043 & 23.91  & 0.00  & 0.0716 & 3.96 \\
12 & 102.446 & 85.381 & 16.66  & 0.00  & 0.0837 & 32.94 \\
15 & 125.051 & 108.226 & 13.45 & 6.07  & 0.2011 & $\geq$600 \\
18 & 146.529 & 133.006 & 9.23 & 16.69 & 0.3237 & $\geq$600 \\
21 & 208.357 & 155.572 & 25.33 & 22.12 & 0.3663 & $\geq$600 \\
\hline
\end{tabular}
\end{table*}

As shown in Table~\ref{tab:steps_time} and Fig~\ref{fig:total_running}, OATH consistently achieves the lowest number of total running steps, indicating that it generates the most efficient allocations and task sequences. More importantly, OATH also achieves the shortest end-to-end execution time across all task sizes. Compared with CBBA, OATH reduces the mission completion time by an average of \textbf{27.4\%}, even though its allocation phase consumes slightly more computation time.  

These results highlight that the overall mission efficiency depends not only on the speed of the allocation process but also, on how well the allocation accounts for the real execution paths. In practice, it is worthwhile to spend additional time in task assignment to obtain higher allocation optimality, since the resulting reduction in execution time outweighs the extra allocation overhead. This confirms that OATH provides the best trade-off between allocation cost and overall mission efficiency.

\nan{To further evaluate the solution quality of OATH, we added a small-scale comparison against an optimization-based MILP baseline solved using Gurobi. 
In this experiment, we compare OATH with the MILP solution in terms of total travel distance and computation time for problem instances with $9{-}21$ tasks. The results are summarized in Table~\ref{tab:steps_time_gap}. 
For small instances ($9$ and $12$ tasks), where the MILP solver terminates with certified optimality, OATH produces feasible solutions with relative distance gaps of $23.91$\% and $16.66$\%, respectively. While OATH does not achieve global optimality in these cases, it computes solutions within $0.07{-}0.08$ seconds, compared to $3.96{-}32.94$ seconds for MILP. This highlights a clear trade-off: OATH sacrifices a portion of optimality in exchange for significantly faster computation.}

\nan{For larger instances ($15$ tasks and above), the MILP solver reaches the imposed $600$-second time limit and does not return a certified optimal solution within the allowed runtime.
To provide context, we report the remaining optimality gap of the MILP solution with respect to the best-known lower bound, which ranges from $6.07$\% to $22.12$\%. This indicates that even the optimization-based baseline becomes computationally impractical in this regime. 
In contrast, OATH consistently produces feasible solutions within sub-second computation time ($0.20{-}0.37$ seconds), while maintaining competitive solution quality relative to the best MILP solutions obtained within the time limit. These results show that OATH provides a good trade-off between solution quality and computation time, which is essential for real-time heterogeneous task allocation.}

\subsection{Results on Scalability Performance}
\nan{To evaluate the scalability of OATH, we conduct two experiments. 
The first experiment examines the MILP solver time, while the second evaluates the overall system performance as the number of tasks increases.}

\begin{figure}[t]
    \centering
    \includegraphics[width=0.9\linewidth]{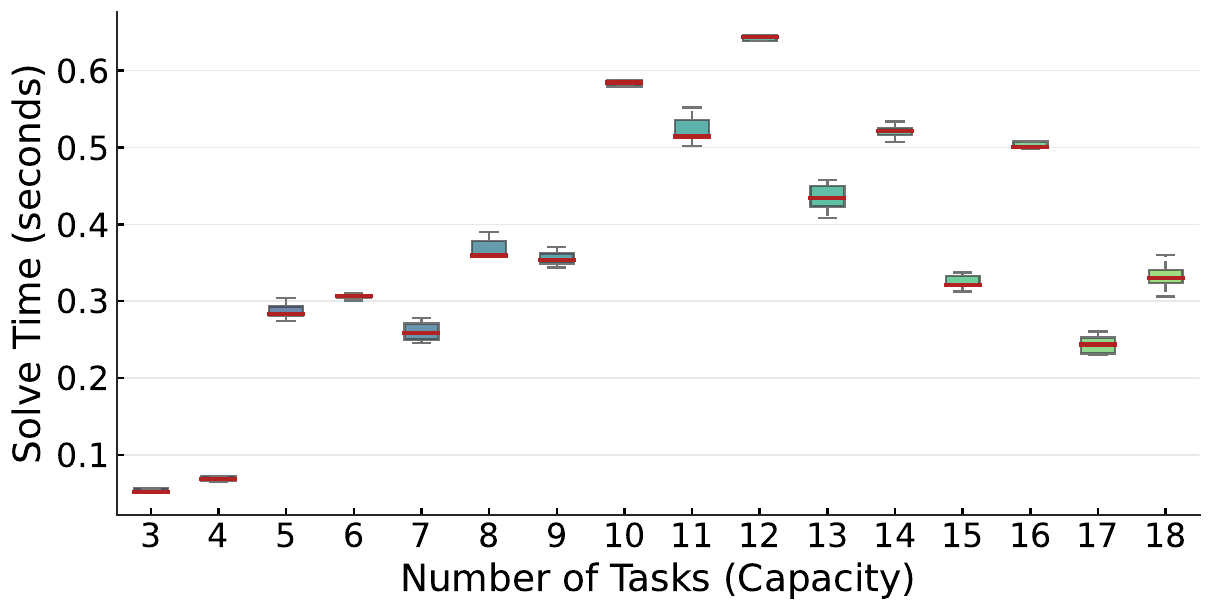}
    \caption{\nan{MILP solver runtime under varying task capacities. 
Boxplots illustrate the distribution of solver runtimes over five independent trials for each task capacity, with the red line indicating the median solve time.}}
    \label{fig:MILP_time}
\end{figure}

\nan{The MILP solver time is an important factor affecting the scalability of OATH.  In the end-to-end experiments, the cluster size does not exceed $10$ tasks per cluster. To further examine the practical scaling behavior of the MILP solver, we additionally evaluate scenarios where a single cluster contains up to $18$ tasks. In this experiment, we gradually increase the number of tasks that the MILP needs to sequence (corresponding to the robot task capacity), and report the solver time over five trials for each setting. The results are shown in Fig.\ref{fig:MILP_time}.}

\nan{As illustrated in Fig.~\ref{fig:MILP_time}, the MILP solver time initially increases as the number of tasks grows, since the problem size and the number of feasible task sequences increase. The solver time reaches its maximum when the number of tasks is around $12$, and remains below $0.7$ seconds in all tested cases. Interestingly, when the number of tasks further increases, the solver time decreases. This is because when the number of tasks to be sequenced approaches the total number of tasks in the cluster, the combinatorial flexibility of the MILP is reduced: most tasks must be included in the route to satisfy the capacity constraint, leaving fewer alternative task subsets and permutations to consider. As a result, the MILP search space becomes more constrained, and the solver effectively focuses on excluding a small number of tasks rather than exploring many competing task sequences, which leads to faster convergence.}

\begin{figure}[t]
  \centering
  \includegraphics[width=\linewidth]{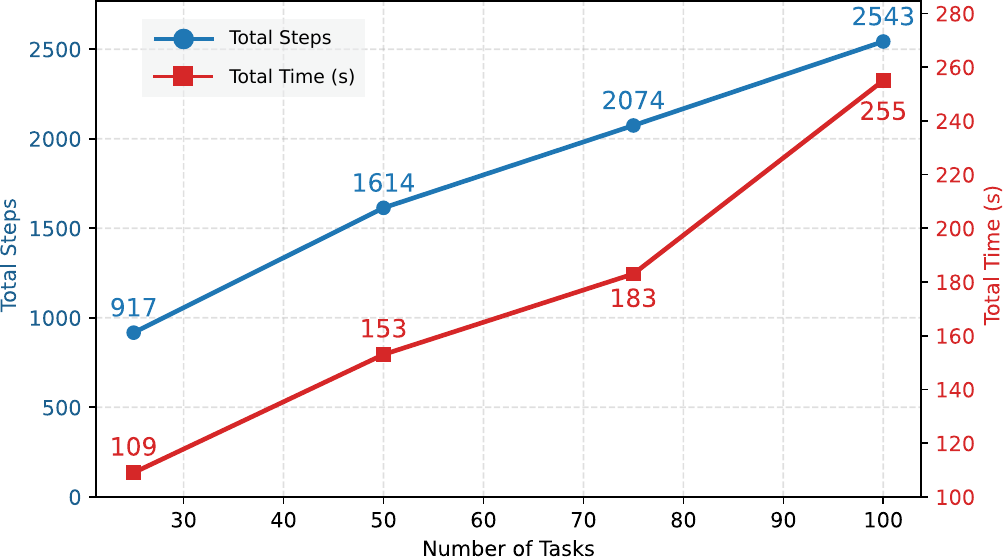}
  \caption{Scalability of \textsc{OATH} with respect to the number of tasks. With the team and planner fixed, increasing $P$ from $25$ to $100$ yields approximately linear growth in \emph{total steps} (left axis) and \emph{total time} (right axis). The near-linear trend and improving per-task cost indicate that \textsc{OATH} scales well to larger workloads.}
  \label{fig:scalability}
\end{figure}

We also assess the scalability of OATH by increasing the number of tasks from $P{=}25$ to $P{=}100$ while keeping the team configuration and planner fixed. As shown in Fig.~\ref{fig:scalability}, both the $T_{\text{total}}$ and $S_{\text{total}}$ grow approximately linearly with $P$, indicating acceptable scaling behavior for online use. Concretely, the total time increases from $109$\,s to $255$\,s ($\times 2.34$) and the total steps from $917$ to $2543$ ($\times 2.77$) as $T$ quadruples ($\times 4$). The amortized cost per task improves with scale: steps per task decrease from $36.7$ to $25.4$ (${-}30.7\%$), and time per task drops from $4.36$\,s to $2.55$\,s (${-}41.5\%$). These trends indicate that OATH exhibits near-linear growth in overall effort while the time and steps per task decrease, demonstrating strong scalability.

\subsection{Results on LLM-Guided Instruction Performance}

\begin{figure}[t]
    \centering
    \includegraphics[width=\linewidth]{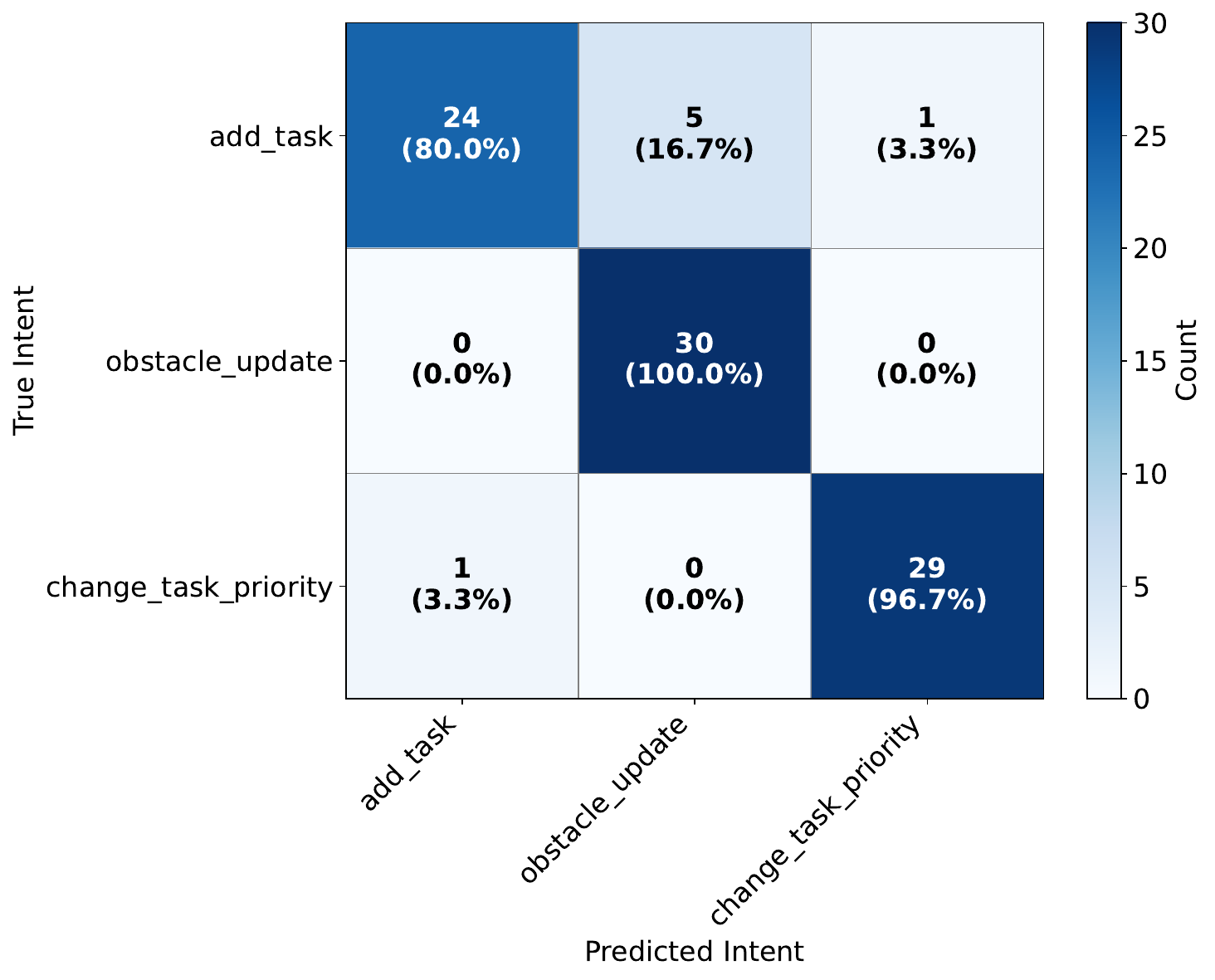}
    \caption{\nan{Confusion matrix for LLM intent parsing.
The matrix reports predicted versus ground-truth instruction intents over 90 test commands, including add new task, obstacle update, and change task priority.
Misclassifications occur only for highly ambiguous inputs, while obstacle update instructions are parsed with $100\%$ accuracy.}}
    \label{fig:confusion_matrix}
\end{figure}

\nan{(i) To evaluate the accuracy of LLM intent parsing, we summarize the parsing results using a confusion matrix, shown in Fig.~\ref{fig:confusion_matrix}.
The parser achieves $100\%$ accuracy for the obstacle update intent.
Misclassifications only occur between the \emph{add new task} and \emph{change task priority} intents.
Specifically, among the \emph{add new task} instructions, five are misclassified as obstacle updates and one is misclassified as a change task priority.
For the \emph{change task priority} intent, one instruction is misinterpreted as adding a new task.
All misclassified cases correspond to highly ambiguous natural-language inputs.
For example, the instruction ``There is a new situation near $(7, 15)$ that needs attention'' is interpreted as an obstacle update because the phrase ``new situation,'' together with the absence of task-specific keywords, suggests a change in the environment rather than the creation of a new task.
Similarly, an instruction such as ``Something else probably needs to be handled now,'' which is labeled as a change task priority, does not specify whether an existing task should be reprioritized or whether a new task should be created.
Due to the lack of explicit semantic cues, the parser interprets this input as adding a new task.}

\nan{Overall, these results show that the LLM parser performs reliably across a wide range of natural-language instructions.
Parsing errors arise primarily when the input is underspecified and lacks clear intent-related cues.
In such cases, the system does not execute the parsed command directly.
Instead, it triggers an interactive clarification step to resolve the ambiguity (see Appendix~\ref{app:llm_ui}). This reduces the risk of incorrect execution and improves system reliability.}

\begin{figure}[t]
  \centering
  \includegraphics[width=\linewidth]{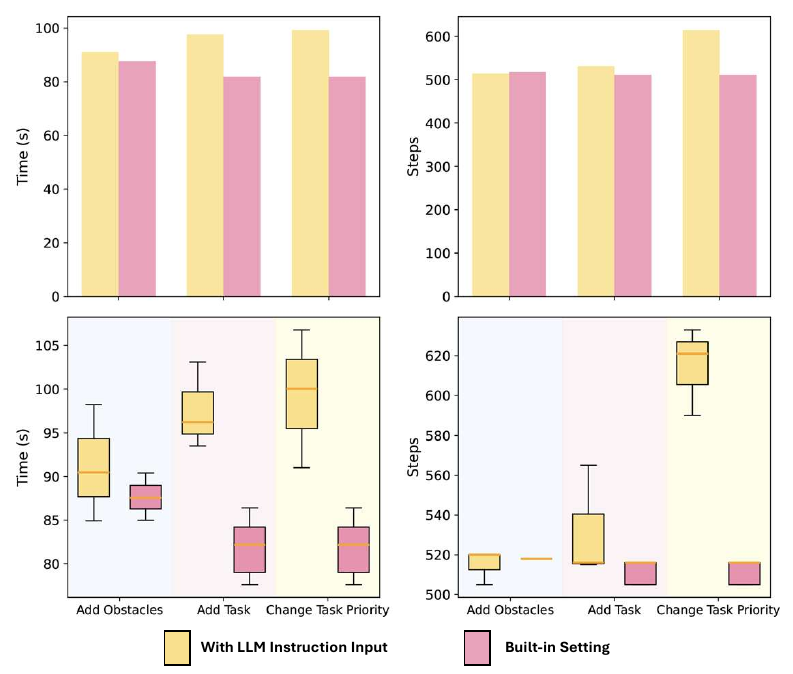}
  \caption{Comparison between \emph{human-issued} and \emph{system built-in} instructions for three instruction types: \textit{New Obstacles Detected}, \textit{Add New Tasks}, and \textit{Change Task Priority}. The left column reports execution time (s), and the right column reports the total number of robot steps. The top row shows means (bar plots), and the bottom row shows variability across trials (box plots). Yellow and pink bars/boxes denote the two modes, respectively, as indicated in the legend.}
  \label{fig:LLM_result}
\end{figure}

(ii) We evaluate LLM-guided instruction performance using two metrics $T_{\text{total}}$ and $S_{\text{total}}$, as shown in Fig.~\ref{fig:LLM_result}. Overall, the LLM-guided settings show larger variance than the system built-in baselines across all instruction types. This variability mainly stems from the timing and context of human interventions, which influence how the system parses instructions and updates the map, task assignment, or execution plan. Specifically:
\begin{enumerate}[label=\alph*)]
    \item \emph{Add New Tasks:} The system built-in baseline completes $25$ tasks efficiently, while the LLM-guided setting (initially $24$ tasks) incurs overhead when the new task is introduced online. The insertion timing strongly influences results: if the new task is added when robots are already near the task region, the additional cost is small; otherwise, it can cause longer detours and higher variance in running time.  
    \item \emph{New Obstacles Detected:} The system built-in baseline achieves stable performance since obstacles are encoded from the beginning. In contrast, the LLM-guided setting shows greater variance in both time and steps. Time variance arises because, when new obstacles appear, robots must pause, check route intersections, and rebuild the product automaton with the updated map. Step variance (the total steps can be smaller than the baseline, as shown in Fig.~\ref{fig:LLM_result}) occurs because if obstacles emerge later, many tasks may already be completed through shortcuts in those regions.
    \item \emph{Change Task Priority:} The system built-in baseline benefits from a fixed, globally optimized priority order. In contrast, the LLM-guided setting must dynamically reorder tasks at execution time. This can force robots to abandon ongoing trajectories and replan, leading to significantly larger variance in both total time and steps, and in some cases noticeably higher overhead than the baseline.  
\end{enumerate}

Overall, these results indicate that while the LLM-guided system enables flexible online adaptation to human instructions, integrating human interaction into MATP introduces a trade-off between adaptability and optimality, as it adds overhead time and increases planning uncertainty.

\section{Hardware Validation}

\nan{To further evaluate the practical applicability of the proposed framework, we carry out real-world experiments using four TurtleBot3 Burger platforms. All experiments are performed fully online, including task allocation, path planning, and LLM-guided instruction processing. No task assignments or trajectories are precomputed offline.}

\begin{figure}[t]
    \centering
    \includegraphics[width=0.9\linewidth]{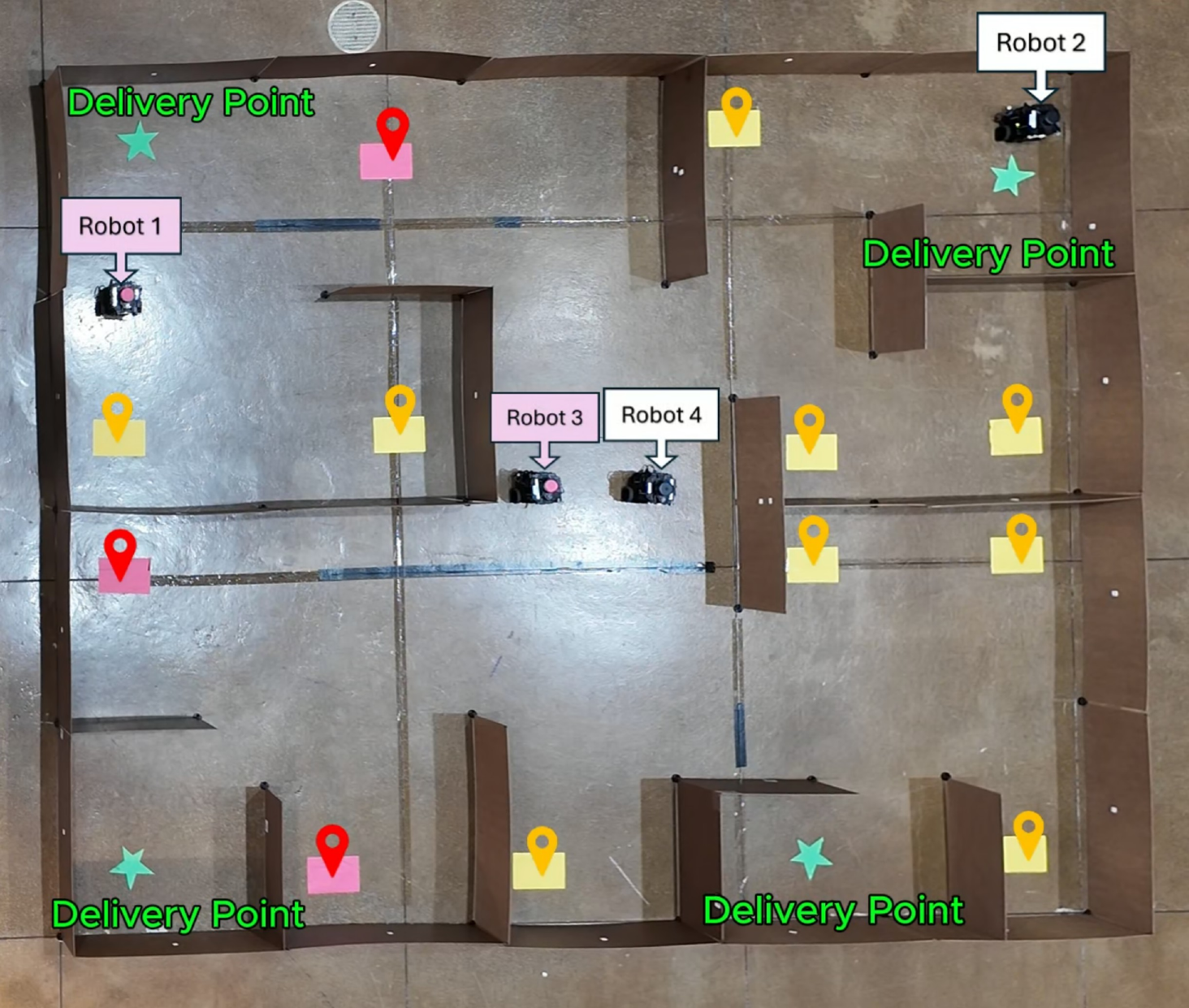}
    \caption{\nan{Hardware experimental setup with four TurtleBot3 Burger robots deployed in a $3.66 \times 4.57$ m indoor environment. Brown panels indicate static obstacles that form corridor structures. Yellow markers represent regular pickup tasks, while pink markers indicate special tasks. Four delivery locations are denoted by green stars. Robots with pink identifiers are capable of executing both task types, whereas the remaining robots are restricted to yellow tasks only. All task allocation and planning are executed online, and robots receive goal positions sequentially from the planner during operation.}}
    \label{fig:Hardware}
\end{figure}

\subsection{Experimental Setup}

\nan{The experiments run in an indoor workspace measuring approximately $3.66 \times 4.57$ m ($12 \times 15$ ft). The environment contains two types of pickup tasks distinguished by color and four delivery locations. In the physical setup, regular tasks are marked in yellow, as shown in Fig.~\ref{fig:Hardware}. Four TurtleBot3 Burger robots are deployed with heterogeneous capabilities. Two robots (identified with pink markers) can execute both task types, while the remaining two robots can only execute the yellow task type. This configuration directly corresponds to the heterogeneous capability matrix used in the planner.}

\nan{All components of OATH operate online within a hybrid centralized–distributed architecture. Obstacle-aware clustering and the cluster-weighted auction (CWA) are executed on a central workstation to determine cluster assignments among robots. After cluster assignment, each robot runs an independent ROS node that communicates with the central planner. The intra-cluster MILP task selection and global path planning are computed by the planner, while each robot maintains its own execution node responsible for receiving goals, monitoring progress, and reporting task completion. The planner dispatches goal positions sequentially to each robot according to its assigned task sequence. Each TurtleBot navigates to the received goal using ROS2 Navigation2. Upon reaching the assigned goal, the robot sends a \texttt{goal\_reached} signal back to the planner, which updates the task state and issues the next target if needed. This closed-loop interaction continues until all tasks are completed.}

\subsection{Hardware Experiment 1: Online Task Allocation and Execution}

\nan{In the first hardware experiment, we evaluate the complete allocation and execution pipeline with four robots and twelve tasks. Each robot possesses a carrying capacity of two tasks. The system successfully performs obstacle-aware clustering, heterogeneous cluster assignment, and intra-cluster sequencing online.}

\nan{Throughout the experiment, robots strictly respect capability constraints. Robots without pink markers never attempt incompatible tasks, while pink-marked robots handle both types when necessary. The planner–robot interaction remains stable during iterative goal dispatch and feedback, and no task synchronization issues are observed. The experiment video is shown in the project website \href{https://llm-oath.github.io/}{https://llm-oath.github.io/}.}

\subsection{Hardware Experiment 2: LLM-Guided Replanning}

\nan{Next, in the second experiment, we validate the LLM-guided planner on the hardware, where human instructions issued during execution are parsed into structured intents that trigger online task reassignment or replanning.}

\nan{In this experiment, two representative instructions are introduced: (1) the detection of a new obstacle and (2) the insertion of a new task. Upon receiving each instruction, the planner updates the task allocation and path plan accordingly in real time. The updated goal positions are dispatched to the corresponding robots without interrupting ongoing execution. After receiving new goals, the robots smoothly switch to their updated target positions using ROS 2 Navigation2. The hardware execution remains stable throughout the replanning process, indicating that the LLM-based instruction module integrates consistently with the real robotic execution stack. The experiment video is available on the project website at \href{https://llm-oath.github.io/}{https://llm-oath.github.io/}.}

\subsection{Robustness to Communication and Hardware Constraints}

\nan{Unlike simulation environments that assume ideal communication, real robot systems introduce network latency, packet variability, and sensing noise. The hardware experiments allow us to examine how these practical factors influence system behavior beyond simulation assumptions.}

\nan{First, the proposed framework does not rely on high-frequency distributed consensus. The cluster-weighted auction is executed centrally and only when a new allocation cycle is triggered. Communication between the planner and robots is limited to goal dispatch and goal-reached signals, which are low-bandwidth and event-driven. In our four-robot experiments, this communication pattern did not lead to observable desynchronization. However, as the number of robots increases, the central workstation must handle a larger number of goal dispatch messages, completion signals, and state updates. In wireless networks, higher traffic density may introduce packet delay or temporary congestion, increasing the time between task completion and subsequent allocation updates. While the event-driven architecture avoids iterative consensus overhead, increased communication delay could slow down reallocation cycles when scaling to many robots. In such scenarios, distributed clustering or localized auction mechanisms could be considered to reduce centralized communication load and maintain scalability.}

\nan{Second, computational complexity remains moderate due to the hierarchical design. Clustering, scoring, and intra-cluster MILP operate on bounded task subsets rather than the full task set. In our hardware experiments, planning steps consistently finish before robots reach their next goals. Nevertheless, in larger environments with denser task distributions or increased robot counts, computation time may grow, potentially increasing the interval between allocation cycles. The current architecture remains practical at the tested scale, but further decentralization may be beneficial for large-scale deployments.}

\nan{Third, sensing noise and kinematic deviations are handled at the control layer by the ROS2 Navigation2 stack, while OATH operates at the task-allocation and waypoint level. Moderate localization noise does not invalidate task feasibility, since allocation decisions rely on estimated path costs rather than instantaneous pose measurements. However, larger localization drift or traversal-time deviations may reduce allocation optimality by altering effective travel costs. Additionally, heterogeneous execution speeds may introduce variability in task completion order. The event-driven reallocation mechanism accommodates such asynchrony naturally: allocation cycles are triggered by task completion rather than synchronized clocks. This prevents global deadlock, although efficiency may decrease if execution variance becomes significant.}

\nan{Finally, compared with simulation, physical robots operate at lower speeds and under real sensing uncertainty. While this often provides additional planning margins, it also introduces variability in traversal time. The hardware experiments suggest that the framework remains functional under moderate real-world uncertainty. At larger scales or under harsher communication and sensing conditions, performance may degrade gradually in terms of allocation optimality or responsiveness, rather than feasibility.}

\nan{Overall, the hardware experiments demonstrate that the OATH framework maintains consistent allocation and replanning behavior under realistic communication conditions and sensing uncertainty at the tested scale. These results provide empirical evidence toward bridging the gap between simulation and real-world deployment, while also highlighting considerations for larger-scale implementations.}

\section{Conclusion}

This work presents OATH, an obstacle-aware task assignment and planning framework for MATP in heterogeneous robot teams.
At its core, the framework introduces an adaptive Halton sequence that adjusts sampling density based on obstacle distribution. In addition, the proposed hierarchical cluster–auction–task selection scheme generalizes to any number of task types and reduces allocation complexity. Together, these components enable scalable and suboptimal task assignment for heterogeneous robot teams operating in obstacle-rich environments.

Beyond task assignment, OATH integrates LLMs as persistent interpreters throughout the execution phase. Unlike prior approaches that employ LLMs only for initialization, our framework continuously leverages LLMs to translate natural language instructions into structured constraints and task updates. This design ensures ongoing adaptability to dynamic human intent, unforeseen obstacles, and mission changes.

Extensive simulations in Isaac Lab demonstrate that OATH achieves superior performance over baseline methods in scalability, responsiveness to environmental changes, and resilience to robot failures, successfully managing up to $100$ tasks with real-time performance.

\nan{In addition to simulation validation, we run fully online hardware experiments using four TurtleBot3 platforms to evaluate real-world feasibility. The results confirm stable task allocation, online replanning, and LLM-guided intervention under realistic communication latency and execution uncertainty, demonstrating that the proposed framework operates reliably beyond simulated environments.}

\nan{In future work, we plan to further investigate learning-based approaches and extend the system to highly heterogeneous teams, including humanoid robots \cite{shamsah2023integrated, ren2025accelerating}.}

\appendices

\section{Asymptotic Complexity Analysis}
\label{app:complexity}

\nan{Let $N$ denote the number of roadmap samples, 
$P=|\mathcal{P}|$ the number of tasks, 
$R=|\mathcal{R}|$ the number of robots, 
and $K$ the number of clusters.}

\subsection*{Time Complexity}

\nan{\textbf{Roadmap construction.}
Planar Delaunay triangulation requires 
$T_{\text{tri}} = \mathcal{O}(N\log N)$ \cite{guibas1985primitives}. 
For planar graphs, $E \le 3N-6 = \mathcal{O}(N)$ \cite{de2008computational}.}

\nan{\textbf{Distance matrix construction.}
Dijkstra with a binary heap runs in 
$\mathcal{O}((|V|+|E|)\log |V|)$ \cite{cormen2022introduction}. 
With $|V|=N$ and $E=\mathcal{O}(N)$, we obtain 
$T_{\text{ssp}} = \mathcal{O}(N\log N)$. 
Executed for $P$ sources,
$T_{\mathcal{M}} = \mathcal{O}(PN\log N)$, 
hence 
$T_{\text{pre}} = \mathcal{O}(PN\log N)$.}

\nan{\textbf{Clustering.}
Agglomerative hierarchical clustering requires 
$T_{\text{cluster}} = \mathcal{O}(P^2\log P)$ 
\cite{cormen2022introduction,mullner2011modern}.}

\nan{\textbf{Auction.}
Scoring and coordination yield 
$T_{\text{CWA}} = \mathcal{O}(IRK)$.}

\nan{\textbf{Intra-cluster MILP.}
For robot $r$ with $n_r$ tasks \cite{graham1994concrete},
\begin{equation}
(2n_r)! 
\approx 
\sqrt{4\pi n_r}
\left(\frac{2n_r}{e}\right)^{2n_r}
\end{equation}}

\nan{Thus 
$T_{\text{MILP},r} = \mathcal{O}(c^{n_r}),\; c>1$, 
consistent with NP-hard routing problems \cite{junger1995traveling}. 
In OATH, clustering constrains $n_r$ to remain small, so the MILP size is structurally bounded in practice.}

\nan{\textbf{Product automaton.}
The state space satisfies 
$|\mathcal{V}_{prod}| = |\mathcal{V}_B|N$  \cite{baier2008principles}.}

\nan{Graph search complexity is \cite{cormen2022introduction,koenig2002d}
\begin{equation}
T_{\text{plan}} =
\mathcal{O}\!\left(
(|\mathcal{V}_{prod}|+|\mathcal{E}_{prod}|)
\log |\mathcal{V}_{prod}|
\right)
\end{equation}}

\subsection*{Overall Time Complexity}

\nan{Initial allocation cycle:
\begin{equation}
\begin{aligned}
T_{\text{init}}
=&\;
\mathcal{O}(PN\log N)
+\mathcal{O}(P^2\log P) \\
&+\mathcal{O}(IRK)
+\sum_{r=1}^{R}\mathcal{O}(c^{n_r}) \\
&+\mathcal{O}\!\left(
(|\mathcal{V}_{prod}|+|\mathcal{E}_{prod}|)
\log |\mathcal{V}_{prod}|
\right).
\end{aligned}
\end{equation}}

\nan{Subsequent cycles:
\begin{equation}
\begin{aligned}
T_{\text{cycle}}
=&\;
\mathcal{O}(P^2\log P)
+\mathcal{O}(IRK) \\
&+\sum_{r=1}^{R}\mathcal{O}(c^{n_r}) \\
&+\mathcal{O}\!\left(
(|\mathcal{V}_{prod}|+|\mathcal{E}_{prod}|)
\log |\mathcal{V}_{prod}|
\right).
\end{aligned}
\end{equation}}

\subsection*{Space Complexity}

\nan{Roadmap storage requires $\mathcal{O}(N)$.
The distance matrix requires $\mathcal{O}(P^2)$.
The product automaton requires $\mathcal{O}(|\mathcal{V}_B|N)$ 
\cite{baier2008principles}.
Total memory:
\begin{equation}
    \mathcal{O}(N + P^2 + |\mathcal{V}_B|N)
\end{equation}}

\section{Interactive Human--LLM Command Interface and Prompt Design}
\label{app:llm_ui}

\nan{To support more natural and less structured user inputs during task execution, we implement an interactive human--LLM command interface that complements the instruction-parsing pipeline described in Sec.~\ref{sec:llm_translator}.
Rather than requiring users to provide fully specified commands in a single utterance, the interface accepts high-level, ambiguous, or incomplete instructions and supports a short clarification dialogue between the user and the LLM before committing any update to the planner.}

\begin{figure}[t]
    \centering

    \includegraphics[width=0.75\linewidth]{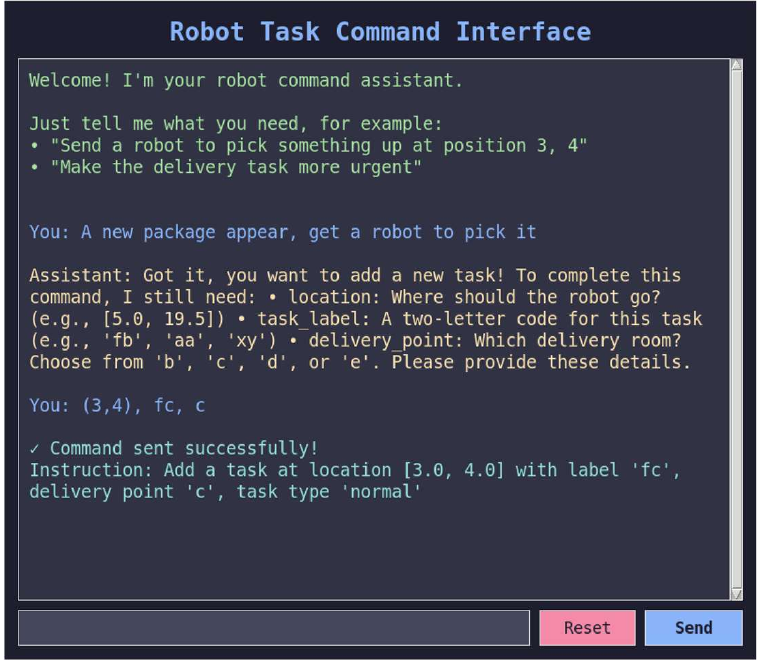}
    \vspace{2mm}

    \includegraphics[width=0.75\linewidth]{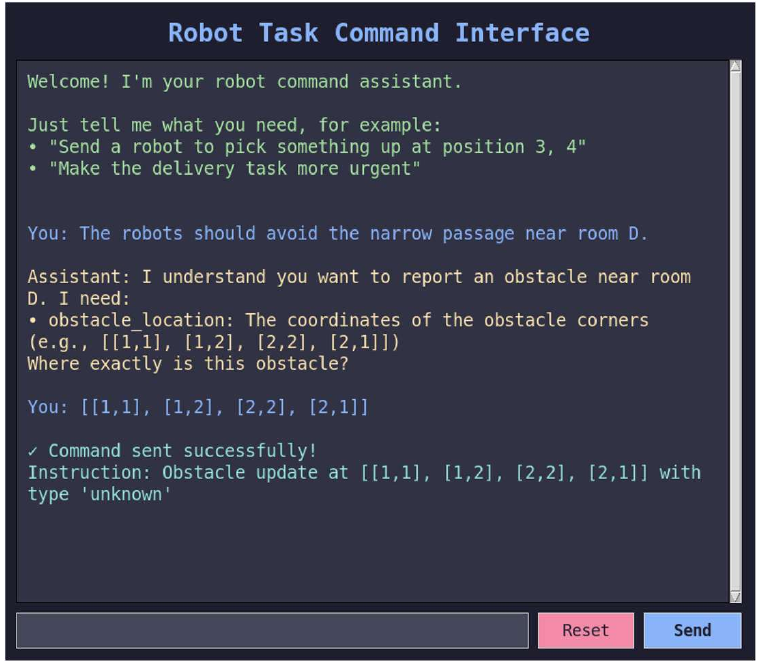}
    \vspace{2mm}

    \includegraphics[width=0.75\linewidth]{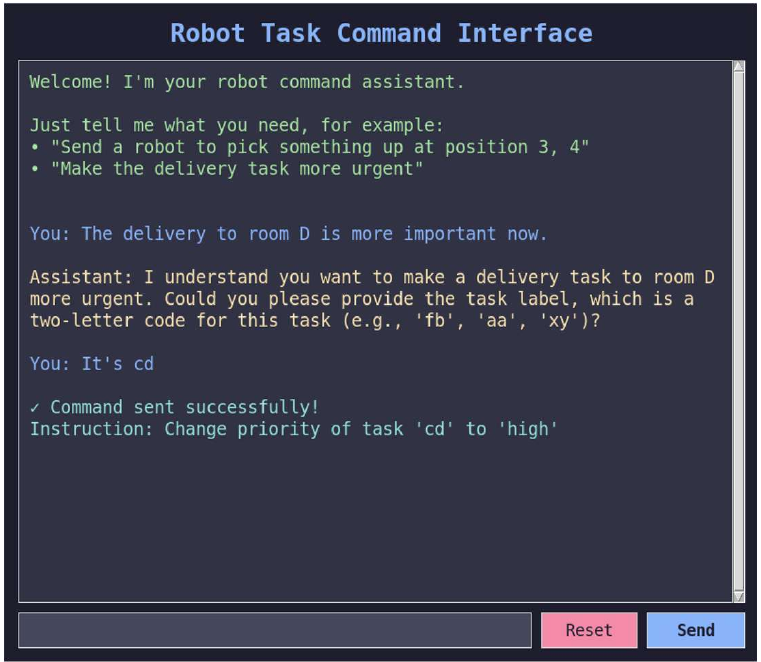}

    \caption{\nan{Interactive human--LLM command interface under three instruction types:
    (top) adding a new task,
    (middle) updating obstacle information,
    and (bottom) changing task priority.
    The interface exposes the LLM's parsed interpretation for human confirmation before execution.}
    }
    \label{fig:llm_ui}
\end{figure}

\nan{The interaction follows a turn-based design, as illustrated in Fig.~\ref{fig:llm_ui}.
When a user issues an instruction, the LLM first infers the underlying intent (for example, adding a task, updating obstacles, or changing task priority) and extracts all parameters that can be reliably determined from the input.
If one or more required parameters are missing or ambiguous, such as task location, delivery target, or obstacle position, the LLM explicitly requests additional information from the user.
Only after all required fields are resolved does the system generate a structured command in the predefined JSON format and dispatch it to the planner.}

\nan{In our implementation, the LLM acts strictly as an instruction parser rather than a planner.
After receiving a human command, it performs intent detection and slot filling to verify whether all required fields (for example, task location, task label, and delivery point) are specified.
If critical information is missing or underspecified, the system does not generate values implicitly.
Instead, it asks targeted follow-up questions to resolve the ambiguity before issuing a structured command to the task allocation module.
An example interaction is shown in Fig.~\ref{fig:llm_ui}.}

\nan{For highly ambiguous commands such as ``get that robot over there to handle it,'' the system correctly identifies that essential grounding information is missing, including which robot is being referenced and what location ``there'' corresponds to.
Because the LLM does not have access to a fully grounded world model or real-time robot state information, it cannot resolve such references autonomously.
This limitation represents a known failure mode.
In these cases, the system relies on user clarification rather than making implicit assumptions.
We intentionally adopt this conservative design choice to avoid incorrect task execution.}

\nan{This design provides three practical advantages.
First, it lowers the cognitive burden on human operators by removing the need to remember rigid command templates.
Second, it improves system robustness by preventing partially specified or ill-formed commands from directly triggering replanning or task reassignment.
Third, the interface explicitly exposes the LLM's parsed interpretation of the instruction to the human commander for confirmation, enabling users to verify correctness before execution and reducing the risk of downstream errors caused by misinterpretation.}

\nan{The LLM is guided by a fixed system prompt that defines its role as an interactive command parser and dialogue manager. The complete prompt used in our experiments is provided below.}

\paragraph{\nan{User Interface Prompt.}}
\nan{This prompt is used by the UI to guide the dialogue with the human commander.}

\begin{promptbox}
You are a friendly and intelligent assistant helping users control robot team. You can understand casual, vague, or informal language and interpret the user's intent.

YOUR CAPABILITIES:
You support three types of robot commands:

1. add_task
   REQUIRED:
   - location: [x, y] coordinates.
   - task_label: a two-letter code defined by the user.
   - delivery_point: one of {'b','c','d','e'}
     'b' -> [6, 3]
     'c' -> [5, 16]
     'd' -> [16, 13]
     'e' -> [16, 5.5]
   OPTIONAL:
   - task_type: {'normal','special'} ('normal' by default)

2. obstacle_update
   REQUIRED:
   - obstacle_location: line segment or polygon.
   OPTIONAL:
   - obstacle_type (e.g., 'wall', 'block', 'unknown')

3. change_task_priority
   REQUIRED:
   - task_label

UNDERSTANDING CASUAL LANGUAGE:
- "go grab something from point A" -> add_task
- "there's something blocking the way near X,Y" -> obstacle_update
- "make task X more urgent" -> change_task_priority

CONVERSATION STYLE:
- Be conversational and friendly.
- If the user is vague, make reasonable guesses and confirm.
- Use context from previous turns.

WHEN INFORMATION IS MISSING:
If the intent is clear but required parameters are missing:
1. State what you understood (intent + known parameters)
2. List all missing required parameters
3. Provide example formats for each missing parameter

RESPONSE FORMAT (JSON ONLY):
{
  "status": "complete" | "need_more_info" | "clarification",
  "intent": "add_task" | "obstacle_update" | "change_task_priority" | null,
  "collected_params": {...},
  "missing_params": [...],
  "message": "response to the user",
  "instruction": "final formatted instruction (only if complete)"
}

When status is "complete":
- add_task:
  "Add a task at location [x,y] with label 'id',
   delivery point 'd', task type 'type'."
- obstacle_update:
  "Obstacle update at [[x1,y1],[x2,y2],...] with type 'type'."
- change_task_priority:
  "Change priority of task 'id' to high."

CRITICAL:
Respond with ONLY a valid JSON object.
Do not include explanations or formatting outside JSON.
\end{promptbox}

\paragraph{\nan{Instruction Parser Prompt.}}
\nan{After the UI receives a complete instruction from the human commander, the instruction parser reformulates the instruction and forwards it to the planner.}

\begin{promptbox}
You are a command parser. Convert the following instruction into a JSON array, where each element represents one intent.
Possible intents: "add_task" or "obstacle_update" or "change_task_priority".

Each element should follow one of the following formats:

For "add_task":
{{
  "intent": "add_task",
  "parameters": {{
    "location": [1.0, 2.0],
    "delivery_point": "b",
    "task_label": "fb",
    "task_type": "special"
  }}
}}

For "obstacle_update":
{{
  "intent": "obstacle_update",
  "parameters": {{
    "obstacle_location": [[4.1, 1.1], [4.1, 2.0], [2.5, 2.0], [2.5, 1.1]],
    "obstacle_type": "wall"
  }}
}}

For "change_task_priority":
{{
  "intent": "change_task_priority",
  "parameters": {{
    "task_label": "fb",
    "priority": "high"
  }}
}}

Instruction: {instruction_text}

Return ONLY the JSON array. Do NOT add any explanation.
\end{promptbox}

\section{LLM-Guided Instruction User Study}
\label{app:user_study}

\begin{figure}[t]
    \centering
    \includegraphics[width=0.9\linewidth]{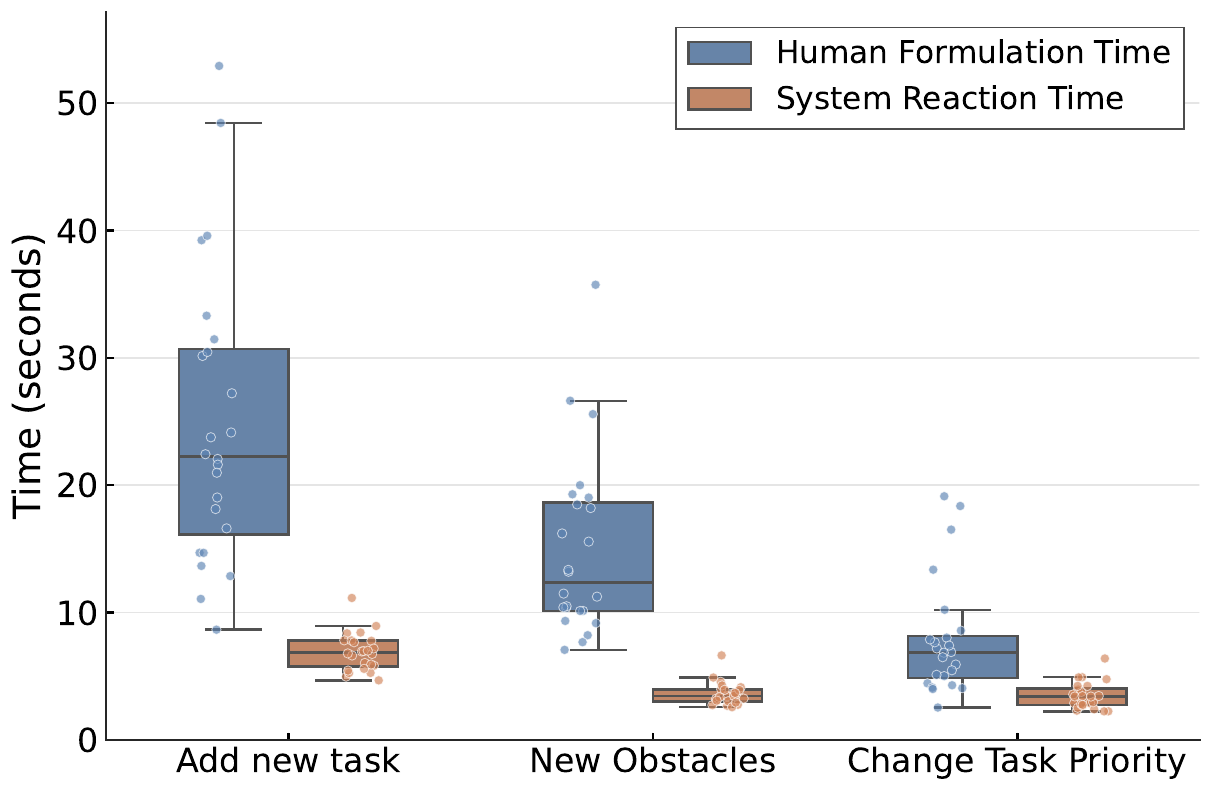}
    \caption{\nan{Comparison between human formulation time and system reaction time for different types of human-in-the-loop interventions. The system reaction time includes both LLM parsing time and the subsequent replanning time. Results are reported for three representative scenarios: adding a new task, introducing new obstacles, and changing task priorities.
    }}
    \label{fig:user_study}
\end{figure}

\begin{figure}[t]
    \centering
    \includegraphics[width=0.9\linewidth]{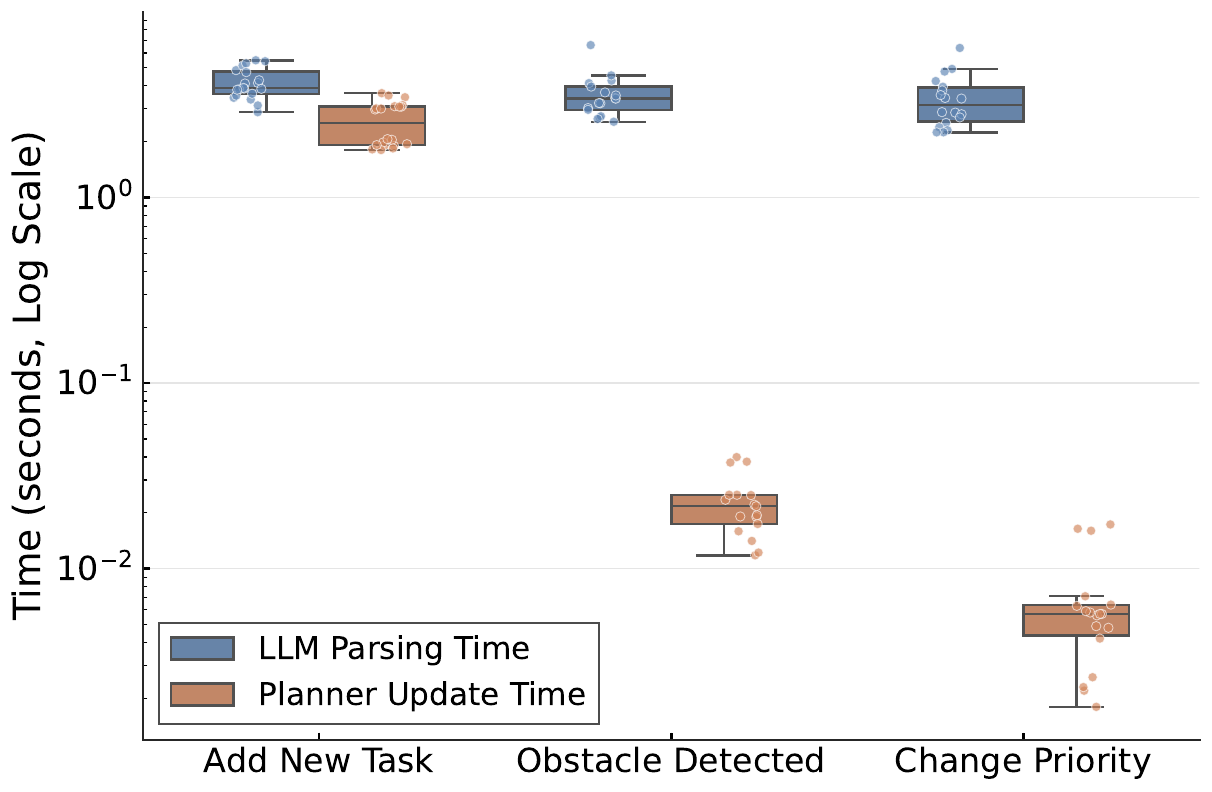}
    \caption{\nan{Breakdown of response latency in LLM-triggered interventions.
    Blue boxplots denote the LLM inference (natural language parsing) time, 
    while orange boxplots denote the planner update time.
    The y-axis is shown in logarithmic scale.
    Across all three intervention types (Add New Task, Obstacle Detected, and Change Priority), 
    the LLM parsing stage dominates the total latency, whereas the replanning module 
    operates at the millisecond level. This separation clarifies that the primary 
    bottleneck arises from model inference rather than from the task allocation algorithms.}}
    \label{fig:LLM_system}
\end{figure}

\nan{In addition to evaluating intent parsing accuracy and system performance impact, we conduct a user study to quantify human-in-the-loop efficiency and analyze the latency of the LLM-guided pipeline.}

\nan{We recruit eight participants to evaluate the efficiency of human interventions. Each participant performs three trials for each intervention intention, resulting in 24 trials per intention. Participants are allowed to freely design their own instructions in each trial. In particular, they may introduce new tasks or new obstacles at arbitrary locations in the environment, without being restricted to predefined templates or fixed coordinates.}

\nan{To evaluate the efficiency and latency characteristics of the LLM-guided pipeline, we conduct two complementary experiments. The first experiment examines the overall human-in-the-loop interaction by measuring human formulation time and total system reaction time. The second experiment further decomposes the system reaction time into LLM inference time and planner update time, allowing us to analyze the source of latency under different intervention types.}

\paragraph{Human Formulation vs. System Reaction Time.}

\nan{The comparison between human formulation time and system reaction time is shown in Fig.~\ref{fig:user_study}. 
Across all three intervention types (\emph{Add New Task}, \emph{New Obstacle Detected}, and \emph{Change Task Priority}), the human formulation time exhibits larger variance than the system reaction time. This variance is expected, as different participants compose instructions at different speeds and with varying levels of detail.}

\nan{Among the three intervention types, \emph{Add New Task} results in the longest human formulation time. This is consistent with the higher information requirement of this intention, which typically involves specifying task location, delivery point, and task type. In contrast, \emph{New Obstacle Detected} and \emph{Change Task Priority} require fewer parameters and therefore yield shorter formulation times.}

\nan{The system reaction time remains relatively stable across participants and intervention types. In all cases, the reaction time is consistently lower and less variable than the human formulation time. This indicates that once a valid instruction is provided, the LLM parsing and replanning pipeline introduces limited additional delay.}

\paragraph{Latency Decomposition.}

\nan{The breakdown of response latency is shown in Fig.~\ref{fig:LLM_system}. The dominant contributor to the total response latency is the LLM inference stage rather than the planner update stage.}

\nan{More specifically, for the \textit{Add New Task} scenario, both the LLM parsing time and the system response (planner update) are on the order of $10^{0}$ seconds. The LLM parsing typically falls in the several-second range (roughly 2--4 s), while the system response time is also seconds-level (roughly 1--3 s). This indicates that adding a new task triggers a substantially heavier planning workload, including inserting a new node into the Halton-based roadmap, updating the task-to-task distance structure, performing task reallocation, and recomputing the corresponding routes, so the end-to-end latency is not dominated by LLM inference alone in this case.}

\nan{In contrast, for the \textit{Obstacle Detected} and \textit{Change Priority} scenario, the system response time drops to approximately $10^{-3}$--$10^{-2}$ seconds (tens of milliseconds), which is about two orders of magnitude smaller than the LLM parsing time.}

\nan{Overall, these results indicate that the primary latency source in the human-in-the-loop setting is the natural language parsing stage rather than the task allocation or replanning algorithms. The planner update remains computationally lightweight relative to the LLM inference overhead.}


\bibliographystyle{IEEEtran}
\bibliography{ref}

\end{document}